%% file: arxiv.tex
\documentclass{article} % For LaTeX2e
\usepackage{iclr2027_conference,times}
\newif\ificlrfinal
\iclrfinalfalse

\usepackage{hyperref}
\usepackage{url}
\usepackage{etoc}

\usepackage{graphicx}
\usepackage[utf8]{inputenc} % allow utf-8 input
\usepackage[T1]{fontenc}    % use 8-bit T1 fonts
\usepackage{booktabs}       % professional-quality tables
\usepackage{amsfonts}       % blackboard math symbols
\usepackage{nicefrac}       % compact symbols for 1/2, etc.
\usepackage{subcaption}     % for subfigure environment
\usepackage{microtype}      % microtypography
\usepackage[dvipsnames]{xcolor}         % colors
\usepackage{amsmath}
\usepackage{tcolorbox}
\usepackage{enumitem}
\usepackage{wrapfig}

\definecolor{ink}{HTML}{1F1426}
\definecolor{deepemerald}{HTML}{008F68}
\definecolor{softgreen}{HTML}{E8F6F1}
\definecolor{creamgold}{HTML}{A97818}

\hypersetup{
    colorlinks=true,
    linkcolor=deepemerald,
    citecolor=creamgold,
    urlcolor=deepemerald
}

\definecolor{smallcolor}{HTML}{3F6FA3}
\definecolor{mediumcolor}{HTML}{4F8A70}
\definecolor{largecolor}{HTML}{B05A7A}

\iclrpreprint

\title{Training Stability under Adaptive Curricula}

\author{%
\makebox[\textwidth][c]{%
Johan Obando-Ceron\textsuperscript{*1,2}\quad
Walter Mayor\textsuperscript{*1,2}\quad
Michael Dennis\textsuperscript{4}
}\\[0.1em]
\makebox[\textwidth][c]{%
\textbf{Aaron Courville}\textsuperscript{1,2,3}\quad
\textbf{Pablo Samuel Castro}\textsuperscript{1,2,4}
}\\[1em]
\makebox[\textwidth][c]{%
\textsuperscript{*}Equal contribution\quad
\textsuperscript{1}Mila -- Qu\'ebec AI Institute\quad
\textsuperscript{2}Universit\'e de Montr\'eal
}\\
\makebox[\textwidth][c]{%
\textsuperscript{3}CIFAR AI Chair\quad
\textsuperscript{4}Google DeepMind
}\\[1em]
\makebox[\textwidth][c]{%
\texttt{\{jobando0730, waltermayor, aaron.courville\}@gmail.com}
}\\[4pt]
\makebox[\textwidth][c]{%
\texttt{\{dennismi,psc\}@google.com}
}
}

\makeatletter
\def\addcontentsline#1#2#3{%
\addtocontents{#1}{\protect\contentsline{#2}{#3}{\thepage}{\@currentHref}{}}}
\makeatother

\begin{document}
\etocdepthtag.toc{mtoc}
\etocsettagdepth{mtoc}{none}
\etocsettagdepth{appendix}{none}

\newbox\dummybox
\setbox\dummybox\vbox{%
  \tableofcontents 
}

\maketitle

\begin{abstract}

Unsupervised Environment Design (UED) approaches train reinforcement learning (RL) agents
on diverse curricula that may evolve with their capabilities. While prior work
has largely focused on how these curricula are constructed, we study whether
their students remain trainable under the resulting distributions. Across
multiple UED mechanisms, environment complexities, and extended training
horizons, we identify recurring learner-side pathologies, including parameter
growth, degraded value--outcome ordering, and loss of representational
diversity, that accompany stagnation and performance degradation. These
findings connect limitations in UED to optimization failures previously
identified in conventional deep RL, while showing how adaptive curricula couple
the learner's optimization state to its future training distribution.
Controlled interventions show that mitigating selected pathologies improves
learning. Stabilizing the learner improves long-horizon performance and
zero-shot generalization without changing the curriculum rule or increasing
environment interaction. Our results establish learner trainability as a
complementary design principle for UED. Effective curricula require students
that remain capable of learning from the distributions they produce.

\end{abstract}

\begin{center}
\begin{minipage}{0.84\linewidth}
\itshape
``The teacher can advance learning only by influencing what the student
does to learn\footnotemark.''

\hfill --- Herbert A. Simon
\end{minipage}
\end{center}

\footnotetext{UED instantiates this principle algorithmically. While prior work has focused
on giving examples a student can effectively learn from, we study whether it sets up the student to
continue to learn reliably in the future.}

\section{Introduction}
\label{sec:intro}

Unsupervised Environment Design (UED) trains deep RL agents
on environment distributions that are constructed and adapted throughout
training to promote generalization to unseen tasks. Rather than relying on a fixed task set, UED methods
generate, select, edit, or replay environments throughout training
\citep{dennis2020emergent,jiang2021prioritized,jiang2021replay,
parkerholder2022evolving,rutherford2024no}. Many of these methods adapt
the training distribution to the evolving capabilities of a student, directing
experience toward tasks estimated to be informative for its current policy.
This literature has therefore largely centered on constructing better
curricula. We study the complementary question of whether the student remains
trainable under the distributions those curricula produce.

Deep RL training is already non-stationary because policy updates alter the
state--action occupancy distribution, while bootstrapped targets depend on
evolving value estimates and, in actor--critic methods, an evolving policy
\citep{sutton2018reinforcement,kumar2021implicit}. Adaptive UED adds a feedback
loop in which changes in the student affect which environments are subsequently
generated, selected, or replayed, coupling learner optimization to its future
data. This interaction may compound pathologies documented in conventional
deep RL, including parameter growth, dormant neurons, feature-rank collapse, and unstable critic
and gradient dynamics
\citep{nikishin2022primacy,lyle2023und,sokar2023dormant,
nauman2024overestimation,mayor2025impact,castanyer2025stable,
obando2026simplicial}. Consequently, learner degradation can affect both how experience is used and
which experience is encountered next. The curriculum may therefore continue
to assign high priority to environments even as the student's capacity to
benefit from them declines. 

\begin{figure*}[!t]
    \centering
    \includegraphics[width=\textwidth]
    {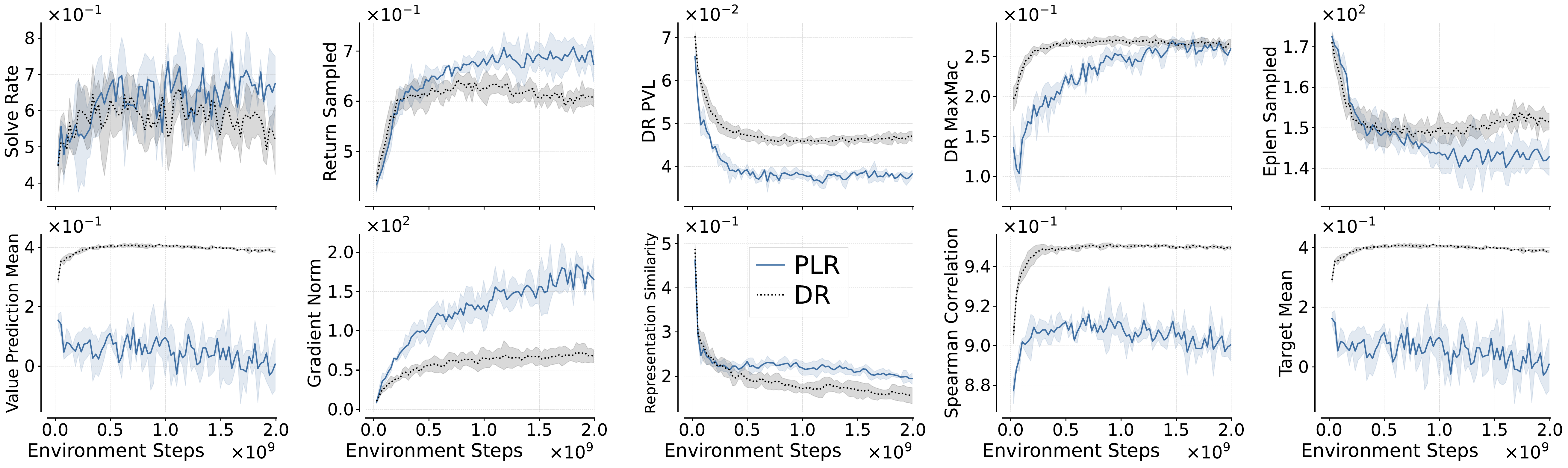}
    \vspace{-0.7cm}
    \caption{\textbf{Adaptive replay reveals a performance--trainability
    tension.}
    Compared with domain randomization (DR; \citet{tobin2017domain}), PLR achieves higher sampled return while exhibiting larger gradient norms, more aligned representations, and weaker value--outcome rank correlation. Differences in value-based learning signals, episode lengths, and value-target distributions further characterize the distinct training experiences under the two methods. We report the interquartile mean after 2B environment steps, aggregated over 5 seeds per method; shaded regions indicate 95\% stratified bootstrap confidence intervals.}
    \vspace{-0.25cm}
    \label{fig:plr_vs_dr}
\end{figure*}

To motivate our investigation, we compare Prioritized Level Replay (PLR, \citep{jiang2021prioritized}), which adapts level selection using student-dependent priorities, with domain randomization (DR, \citet{tobin2017domain}), which samples levels from a fixed distribution. As shown in \autoref{fig:plr_vs_dr}, PLR achieves higher sampled return while exhibiting larger gradient norms, greater representation similarity, and weaker value--outcome rank correlation. Differences in value-based statistics (PVL and MaxMC, see \autoref{app:dr_pvl}), episode lengths, and value targets further characterize the distinct training experiences under the two methods. Although these observations do not establish that adaptive replay causes learner degradation, they show that improved performance can coexist with less favorable value-ordering and representation diagnostics, alongside different gradient dynamics. This motivates examining how optimization, value prediction, and representations evolve under adaptive curricula, and whether learner-side stabilization can improve sustained learning.

\begin{wrapfigure}[14]{r}{0.50\columnwidth}
    \centering
    \vspace{-0.5cm}
    \includegraphics[width=\linewidth]
    {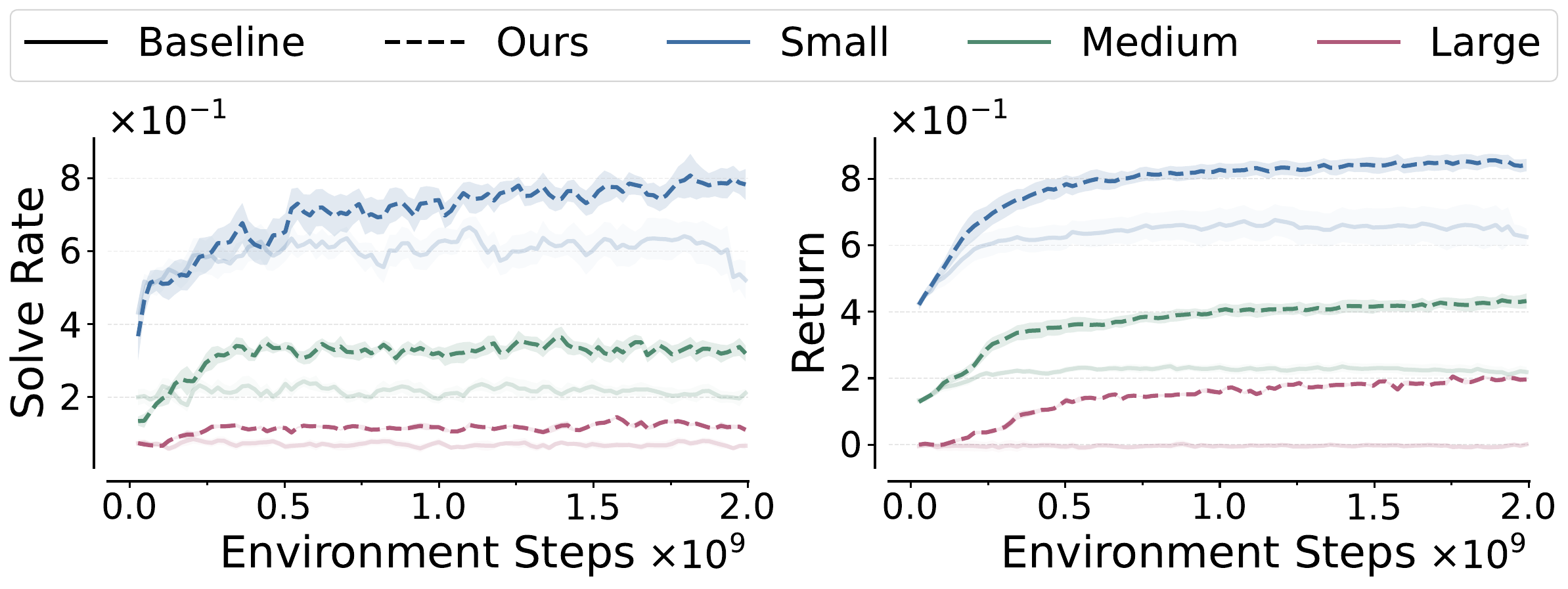}
    \vspace{-0.75cm}
    \caption{\textbf{Learner stabilization improves performance across
        adaptive UED mechanisms.}
        Evaluation solve rate and sampled return aggregated over PLR, SFL, ACCEL,
        and random seeds for
        \textbf{\textcolor{smallcolor}{Small}},
        \textbf{\textcolor{mediumcolor}{Medium}}, and
        \textbf{\textcolor{largecolor}{Large}} Kinetix configurations.
        Faint solid curves denote baselines; dashed curves denote stabilized learners.}
    \label{fig:stabilization_summary}
    \vspace{-0.4cm}
\end{wrapfigure}

We investigate these questions in Kinetix
\citep{matthewskinetix}, whose heterogeneous physics tasks, graded complexity
configurations, and long training horizons provide a demanding testbed.
A detailed analysis of PLR (\autoref{sec:diagnosis}) reveals multiple indicators of declining learner
trainability, including degradation in parameter dynamics, value prediction,
and representations. These indicators become more pronounced in more complex configurations
and over longer training horizons. Complementary to recent work sustaining PPO
learning in Kinetix through massive rollout parallelism
\citep{beukman2026preventinglearningstagnationppo}, we examine learner-side
degradation and whether it can be mitigated without increasing rollout scale.

We evaluate established interventions targeting different aspects of learner
optimization\footnote{Unless stated otherwise, experiments use two billion environment
interactions and five random seeds. Curves report the interquartile mean (IQM) \citep{agarwal2021deep},
and shaded regions show 95\% bootstrap confidence intervals, stratified by
algorithm when results are aggregated.}. Their mixed outcomes show that not all stabilization interventions are equally effective. Weight clipping and categorical value prediction
provide the most consistent individual gains and act at complementary points in
the learning process by controlling parameter scale and reducing the sensitivity
of the value objective to scalar target errors (\autoref{sec:stabilization}). As summarized in
\autoref{fig:stabilization_summary}, the combined intervention improves
performance across multiple curriculum mechanisms, including PLR, SFL \citep{rutherford2024no}, and ACCEL \citep{parkerholder2022evolving}, and across environment complexity levels, without
changing the curriculum rules or environment-interaction budgets. Its benefits
persist over extended training horizons, under limited rollout parallelism,
across sequential distribution shifts, and in zero-shot generalization (\autoref{sec:scaling}). These
results show that learner-side degradation can prevent existing curricula from
realizing their full benefit. Progress in UED therefore depends not only on
constructing informative training distributions, but also on keeping students
trainable under the distributions they produce.

\section{Problem Setting}
\label{sec:setting}

\paragraph{Actor--critic learning.}
We consider actor--critic agents that jointly learn a policy
$\pi_\theta(a\mid s)$ and value approximation $V_\phi(s)$. Our agents use
PPO~\citep{schulman2017proximal}, whose clipped surrogate objective limits the
influence of large likelihood-ratio changes within each optimization phase,
while the critic is trained against trajectory-based targets $\widehat V_t$
using
$\mathcal{L}_V(\phi)=
\mathbb{E}_t[(V_\phi(s_t)-\widehat V_t)^2]$.
Training nevertheless remains non-stationary because policy updates alter the
experience distribution, while newly collected trajectories and evolving value
estimates change the targets and advantages used in subsequent updates. PPO
clipping does not stabilize these quantities or the network representations
across optimization phases. Prior work shows that these dynamics can be
accompanied by representation degradation, loss of plasticity, and learning
stagnation even in on-policy agents
\citep{moalla2024no,juliani2024study,mayor2025impact,
beukman2026preventinglearningstagnationppo}.
See \autoref{app:deep_rl} for details.

\paragraph{Adaptive environment curricula and student coupling.}
UED extends the MDP setting to a level space $\mathcal{L}$, where each level
$\ell\in\mathcal{L}$ defines an environment $\mathcal{M}_\ell$
\citep{dennis2020emergent}. At iteration $k$, the curriculum induces a
distribution $p_k(\ell)$ over levels, and the resulting state--action
distribution is
$d_k(s,a)=\mathbb{E}_{\ell\sim p_k}
[d_{\mathcal{M}_\ell}^{\pi_{\theta_k}}(s,a)]$,
where $d_{\mathcal{M}_\ell}^{\pi_{\theta_k}}$ denotes the occupancy measure
induced by the current policy in environment $\mathcal{M}_\ell$. A training
batch $\mathcal{D}_k$ is collected by sampling levels from $p_k$ and rolling
out $\pi_{\theta_k}$. Let $w_k=(\theta_k,\phi_k)$ denote the student
parameters. The student update is written as
$w_{k+1}=\mathcal{U}(w_k;\mathcal{D}_k)$, where $\mathcal{U}$ denotes the
learning update, while the curriculum evolves according to
$p_{k+1}=\mathcal{C}(p_k,\mathcal{H}_{k+1})$, where $\mathcal{C}$ is the
curriculum update rule and $\mathcal{H}_{k+1}$ is the training history,
including returns, success estimates, or replay statistics generated by the
student. Under domain randomization, $p_k$ remains fixed to the generator
distribution, although $d_k$ still changes as the policy evolves. Adaptive
UED methods instead modify $p_k$ by generating, selecting, editing, or
replaying levels using student-dependent signals
\citep{jiang2021prioritized,jiang2021replay,
parkerholder2022evolving,rutherford2024no}. The curriculum therefore
determines the student's current data, while the updated student influences
its future training distribution. A fixed curriculum rule can consequently
produce different training distributions when coupled to different learners. See \autoref{app:ued} for details.

\section{Diagnosing Optimization Instability in UED}
\label{sec:diagnosis}

\autoref{fig:plr_vs_dr} reveals a performance--trainability tension.
Prioritized Level Replay (PLR) outperforms domain randomization (DR), but also
exhibits larger gradient norms, more aligned representations, and weaker
value--outcome correspondence. This comparison does not establish that adaptive
replay causes learner degradation, but it exposes signs of optimization strain
that performance alone does not reveal. We focus our detailed diagnosis on
PLR~\citep{jiang2021replay,jiang2021prioritized} for two reasons. First, PLR is
a foundational and widely used adaptive UED method. Second, it provides a
particularly transparent view of curriculum--student coupling. Its generator
proposes levels independently of the student, while explicit student-dependent
scores determine which stored levels are replayed. We can therefore examine
how replay priorities, the realized training distribution, and learner
optimization co-evolve without the additional mechanism of explicit level
editing. We later test whether the benefits of stabilization extend to SFL and ACCEL.
SFL uses learnability-based selection, whereas ACCEL builds on PLR's regret-based level prioritization while additionally incorporating level editing to generate new levels through mutations.

\paragraph{Benchmark and evaluation protocol.}
We conduct this analysis in Kinetix~\citep{matthewskinetix}, a heterogeneous
space of procedurally generated physics-based control tasks that supports
billions of environment interactions and provides held-out human-designed tasks
for evaluating zero-shot generalization. Its
\textbf{\textcolor{smallcolor}{Small (S)}},
\textbf{\textcolor{mediumcolor}{Medium (M)}}, and
\textbf{\textcolor{largecolor}{Large (L)}} configurations vary the permitted
numbers of shapes, joints, and thrusters, providing an interpretable axis of
structural complexity. This combination enables us to study optimization over
extended training, its relationship with task complexity, and whether
stabilization transfers beyond the training distribution. 

\begin{figure*}[!t]
    \centering
    \includegraphics[width=\textwidth]
    {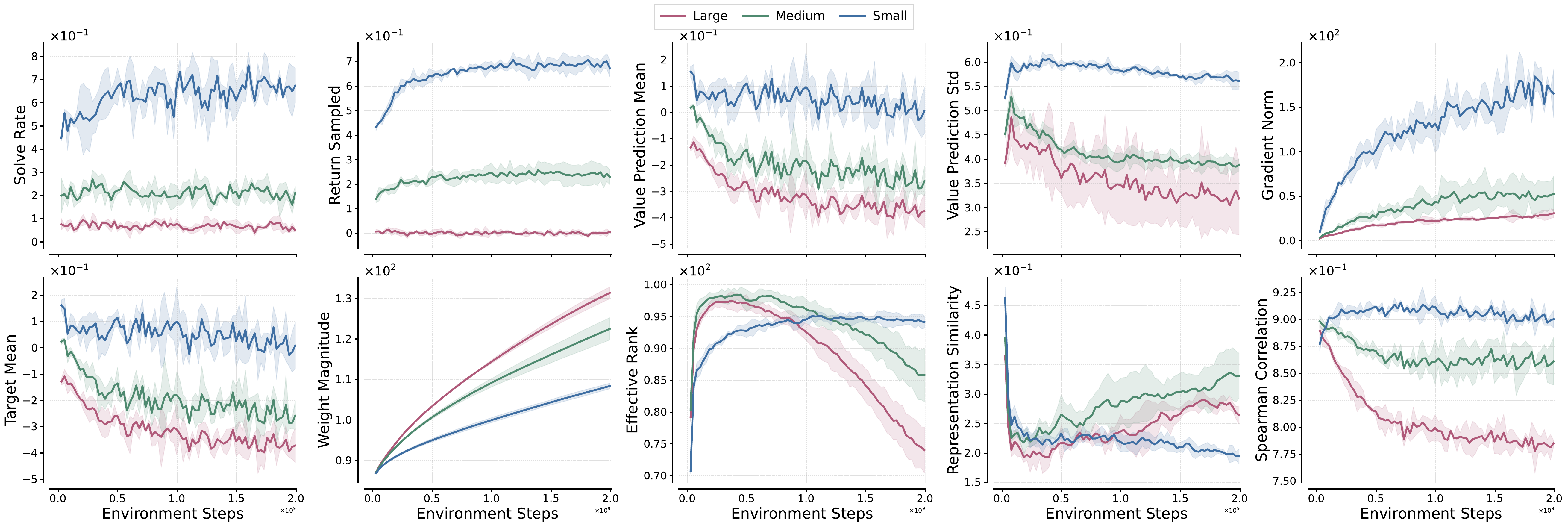}
    \vspace{-0.7cm}
    \caption{\textbf{Performance and learner diagnostics across
    environment complexity.} PLR exhibits weaker performance, greater
    parameter growth, diminishing relative updates, degraded value ordering,
    and loss of representational diversity in more complex Kinetix
    configurations.}
    \vspace{-0.25cm}
    \label{fig:training_pathologies}
\end{figure*}

\paragraph{Optimization and representation diagnostics.}
\autoref{fig:training_pathologies} relates performance to the learner's
optimization, value predictions, and representations. Performance decreases
with environment complexity, while parameter magnitudes increase across
all configurations, with the largest growth in
\textcolor{largecolor}{Large}. Gradient norms also increase across
configurations but are largest in \textcolor{smallcolor}{Small},
indicating that gradient magnitude alone does not explain the
performance ordering.

The critic exhibits a clearer association with deteriorating performance. As complexity increases, value predictions
become more negative and less dispersed, although these statistics also reflect changes in the corresponding target distributions. More importantly, the rank correlation between value predictions and
empirical returns deteriorates most strongly in \textcolor{largecolor}{Large} environments, indicating
weaker value ordering. Representation quality follows a similar pattern. Effective rank declines most
sharply in \textcolor{largecolor}{Large}, while late-training cosine similarity increases,
indicating more aligned and less diverse features. Poor performance is
therefore associated with greater parameter scale, smaller relative updates,
weaker value ordering, and reduced representational diversity. These
diagnostics establish a coherent pattern of degradation, but do not identify a
single cause. Definitions, measured network components, probe distributions,
and aggregation procedures are provided in
App.~\ref{app:metrics}.

\begin{wrapfigure}[16]{r}{0.52\columnwidth}
    \centering
    \vspace{-0.4cm}
    \includegraphics[width=\linewidth]
    {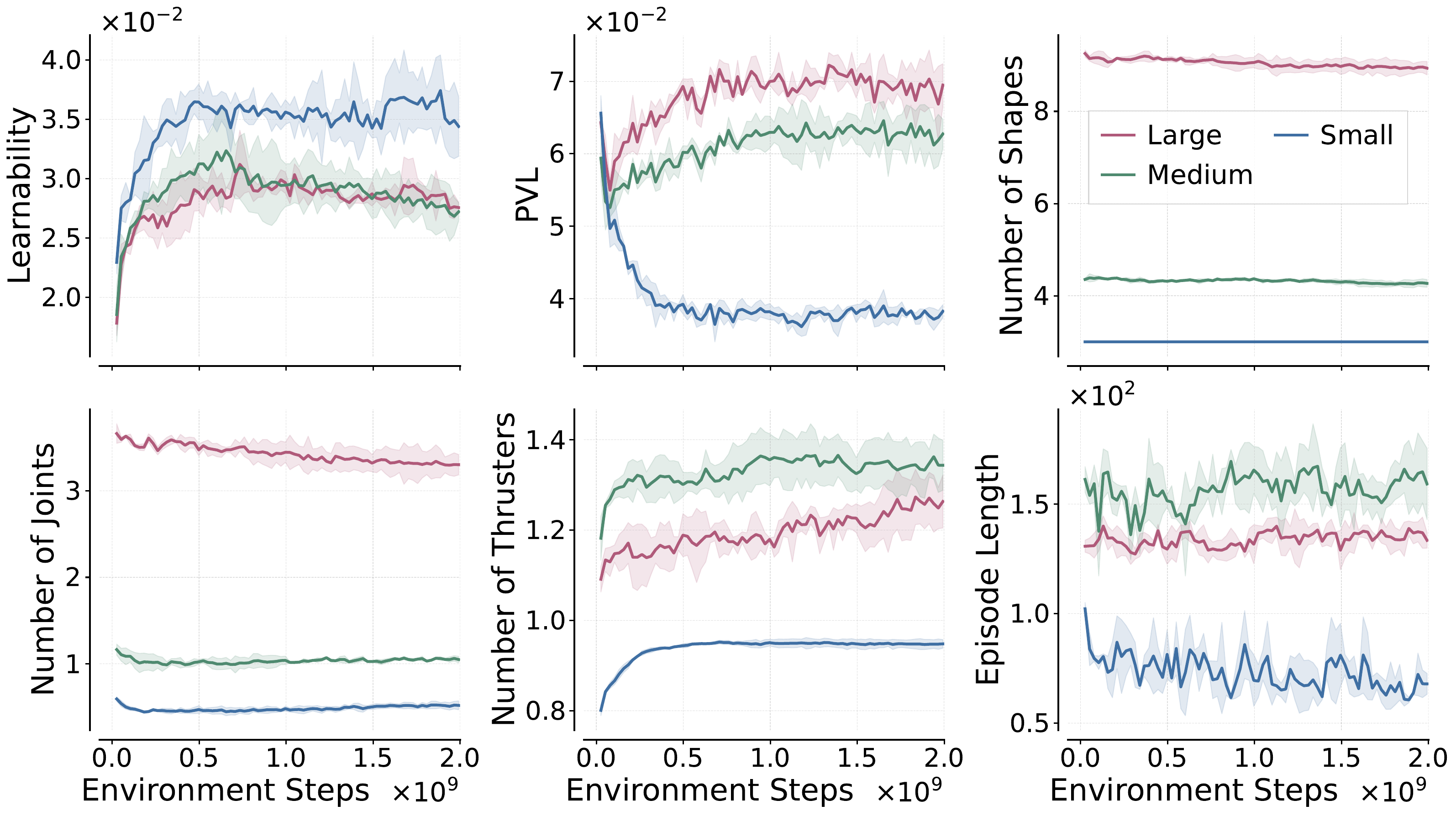}
    \vspace{-0.6cm}
    \caption{\textbf{Curriculum and environment dynamics under PLR.}
    Results for \textbf{\textcolor{smallcolor}{Small}}, \textbf{\textcolor{mediumcolor}{Medium}}, and
    \textbf{\textcolor{largecolor}{Large}} Kinetix configurations. Shaded regions denote 95\% bootstrap confidence intervals across seeds.}
    \label{fig:curriculum_dynamics}
    \vspace{-0.2cm}
\end{wrapfigure}

\paragraph{Curriculum dynamics under PLR.}
To compare PLR's replay priorities with the student's current ability to learn
from the selected levels, we report two student-dependent signals in
\autoref{fig:curriculum_dynamics}. Following
\citet{rutherford2024no}, we define the learnability of level $\ell$ as
$\mathcal{L}_{\mathrm{learn}}(\ell)=p_\ell(1-p_\ell)$, where $p_\ell$ is the
current policy's estimated probability of solving that level. Learnability is
maximal at $p_\ell=0.5$ and vanishes when a level is never or always solved,
making it a proxy for whether the level lies near the student's learning
frontier rather than a direct measure of future improvement. On the other hand, Positive Value Loss (PVL) provides a proxy for regret by accumulating positive value-prediction errors, which PLR uses to prioritize levels for replay.
\citep{jiang2021prioritized,jiang2021replay}.

Mean learnability rises in all configurations but remains highest for Small
environments. PVL follows a different pattern, remaining elevated for Medium
and especially \textcolor{largecolor}{Large} environments. High replay priority (i.e., high PVL) therefore does not
necessarily imply that the selected levels are currently learnable. This
mismatch is clearest in the Large configuration, where PLR continues to assign
high priority despite low learnability and solve rate.

This separation between priority and learnability resembles the
\emph{regret stagnation} phenomenon identified by
\citet{beukman2026preventinglearningstagnationppo}. In that setting, minimax-regret curricula continue
sampling levels with high irreducible regret after the student can no longer
improve on them. We do not establish irreducible regret or minimax-regret
convergence in Kinetix. Instead, we identify a related empirical mismatch and
investigate a complementary learner-side explanation: optimization degradation
may reduce the student's ability to benefit from levels that remain highly
prioritized.

\begin{tcolorbox}[
    colback=softgreen,
    rightrule=0.5mm,
    leftrule=0.5mm,
    top=0.5mm,
    bottom=0.5mm
]
\paragraph{Takeaway.}
At greater complexity, PLR maintains high replay priority despite lower
measured learnability and broader optimization degradation. These associations
do not establish causality, motivating controlled interventions under the same
curriculum rule.
\end{tcolorbox}

\section{Testing the Actionability of Optimization Pathologies}
\label{sec:stabilization}

The preceding analysis shows that parameter growth, weaker value--outcome
correspondence, and deteriorating representations accompany poor performance.
Because these patterns may contribute to degradation or merely reflect it, we
test whether directly targeting them improves learning. We evaluate controlled learner-side interventions while holding the
curriculum rule, actor objective, architecture, interaction budget, and
other training choices fixed.

\subsection{Controlled Optimization Interventions}
\label{sec:interventions}

We screen established interventions designed to control parameter scale,
reduce sensitivity to non-stationary value targets, or preserve representation
quality \citep{juliani2024study,moalla2024no,mayor2025impact,castanyer2025stable}. Our goal is not to identify an
exhaustive or globally optimal stabilization recipe, but to test whether
specific pathologies identified in Sec.~\ref{sec:diagnosis} can be mitigated.
Definitions, hyperparameters, and selection procedures are provided in
App.~\ref{app:interventions}.

\begin{figure*}[!h]
    \centering
    \includegraphics[width=0.28\textwidth]
    {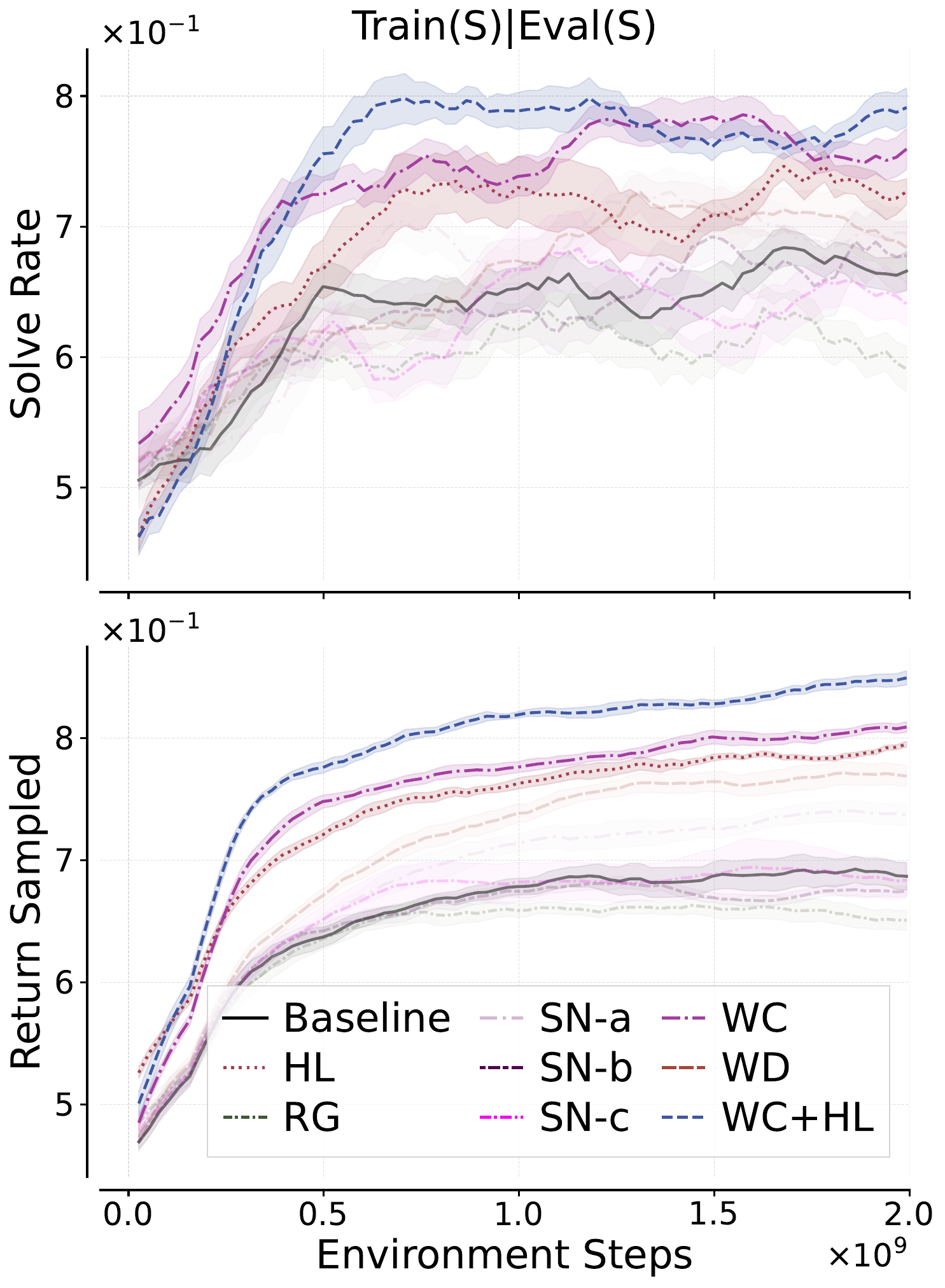}%
    \hspace{0.01\textwidth}
    {\color{black}\vrule width 0.6pt height 5.4cm}
    \hspace{0.01\textwidth}
    \includegraphics[width=0.68\textwidth]
    {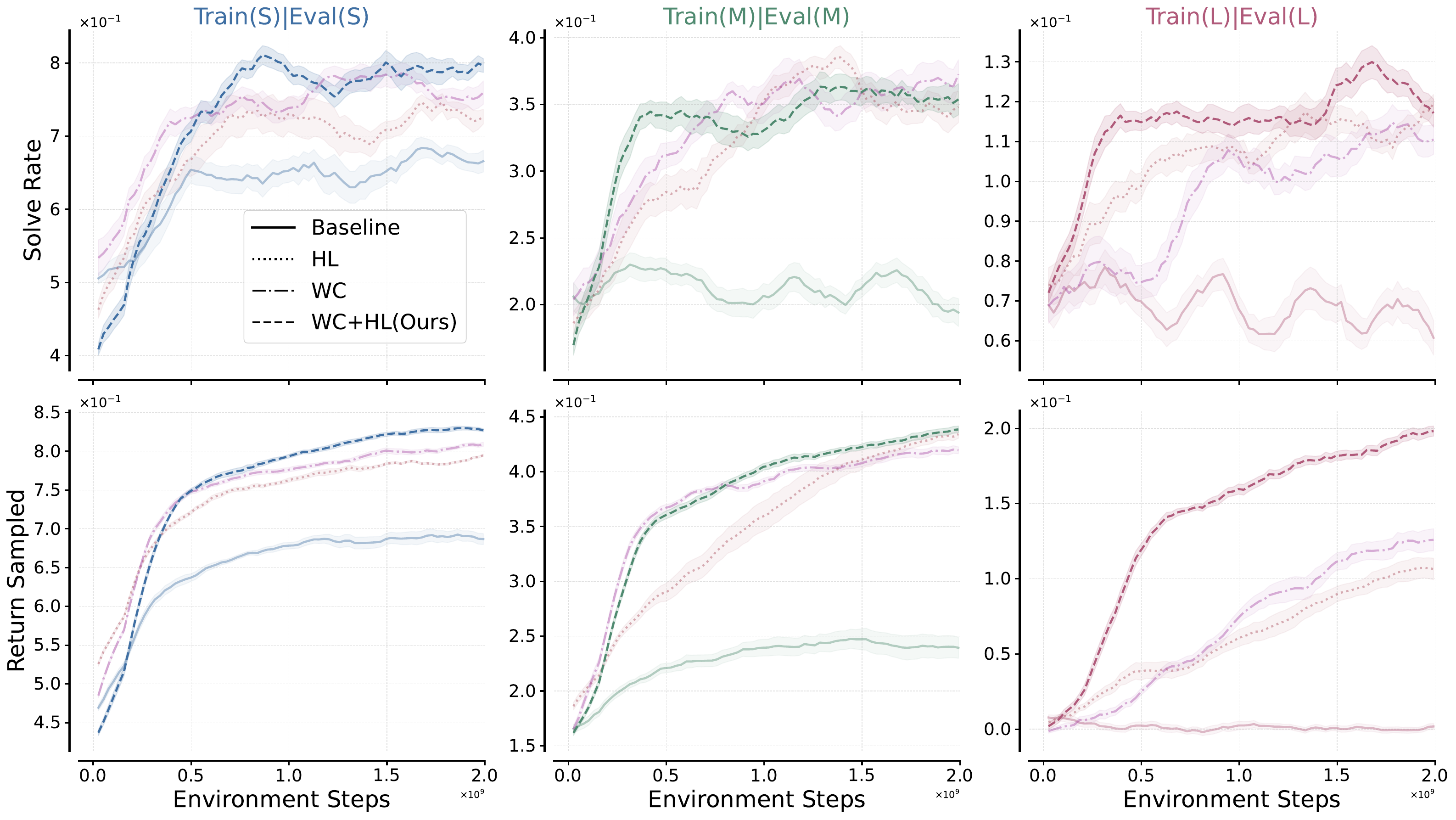}

    \vspace{-0.25cm}
    \caption{\textbf{Intervention screen and factorial ablation under PLR.}
    \textbf{Left:} Evaluation solve rate and sampled return on Small for the
    complete intervention suite. See \autoref{app:acron} for acronym definitions and \autoref{app:metrics} for a detailed explanation. \textbf{Right:} Comparison of the baseline, weight clipping (WC), HL-Gauss (HL), and WC+HL
    across \textcolor{smallcolor}{Small}, \textcolor{mediumcolor}{Medium}, and \textcolor{largecolor}{Large}. Several interventions improve over
    baseline, but WC+HL provides the most consistent gains, including a
    substantially larger return improvement in the most complex Kinetix configuration (\textcolor{largecolor}{Large}).}
    \label{fig:intervention_screen}
\end{figure*}

\autoref{fig:intervention_screen} shows that optimization
interventions differ substantially in their ability to improve
learning under PLR. In the initial screen on
\textcolor{smallcolor}{Small}, weight clipping achieves the
strongest performance among the parameter-control interventions,
while weight decay and HL-Gauss also provide improvements over
baseline. Other interventions yield smaller or less consistent
gains, indicating that regularization alone does not guarantee
better learning. We focus on WC and HL because they perform
strongly while addressing distinct aspects of the learner
degradation identified in Sec.~\ref{sec:diagnosis}. WC explicitly
constrains parameter magnitudes, directly targeting the
parameter growth observed during training. HL-Gauss instead
replaces scalar value regression with categorical prediction
over a discretized value support, providing an alternative
formulation of value learning that can reduce sensitivity to
large scalar target errors. Together, these interventions
combine explicit control of parameter scale with a categorical
value-learning objective
\citep{elsayed2024weight,farebrother2024stop}.
Mathematical definitions and implementation details are provided
in App.~\ref{app:selected}.

The factorial comparison reveals that the contributions of
WC and HL vary with environment complexity. In
\textcolor{smallcolor}{Small}, both interventions improve
performance individually, while WC+HL achieves the highest
final sampled return and strong solve rates. In
\textcolor{mediumcolor}{Medium}, both interventions again
outperform baseline, with WC and WC+HL reaching comparable
final performance. The contrast is most pronounced in
\textcolor{largecolor}{Large}, where WC produces a substantial
improvement in sampled return, while HL alone provides more
modest gains. WC+HL achieves similarly strong performance
to WC, substantially exceeding both the baseline and HL
alone. These findings suggest that controlling parameter
growth becomes particularly relevant in the more challenging
configurations, consistent with the parameter-scale differences
observed in our earlier diagnosis. However, the results do
not establish parameter growth as the direct cause of
performance degradation, nor do they demonstrate a consistent
synergistic benefit from combining the two interventions.

The ablation identifies WC as an effective intervention
across complexity levels, with particularly large improvements
in the most challenging configuration. Although adding HL
does not consistently improve upon WC alone, WC+HL maintains
strong performance across Small, Medium, and Large without
exhibiting substantial degradation relative to the stronger
individual intervention. We therefore select WC+HL as a
practical stabilization configuration for subsequent
experiments, rather than claiming it is globally optimal.
This choice allows us to investigate whether the observed
performance gains are accompanied by improvements in the
learner diagnostics identified earlier, and whether these
benefits generalize beyond PLR.

 \begin{figure*}[!t]
    \centering
    \includegraphics[width=\textwidth]
    {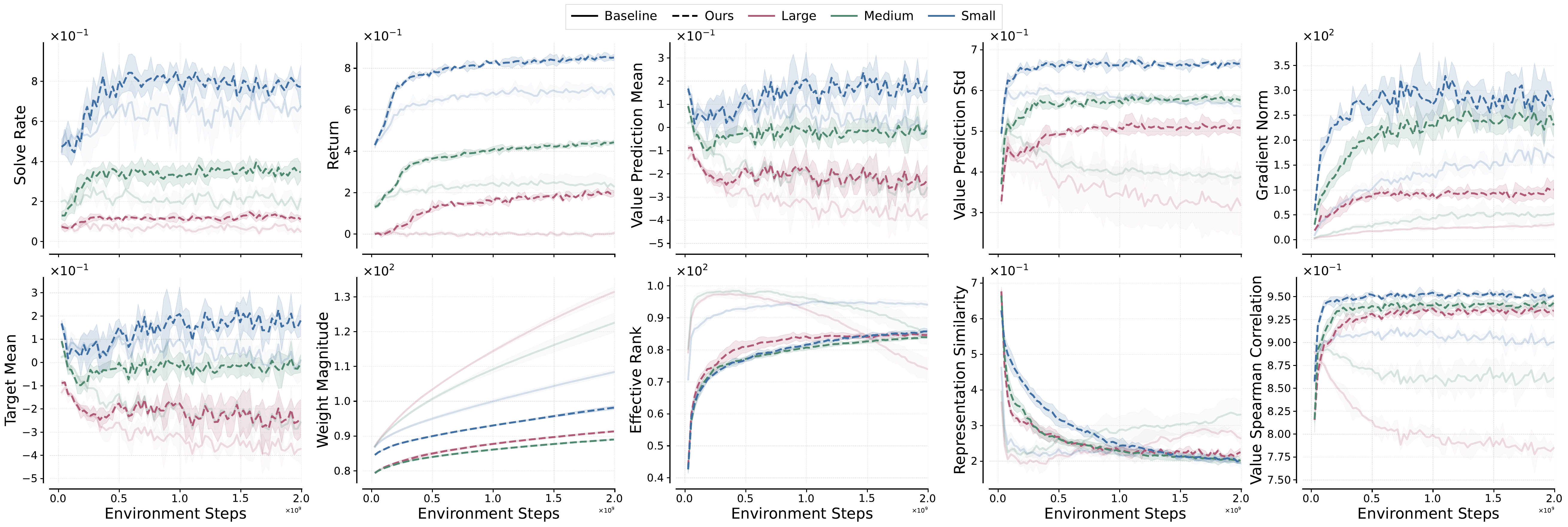}
    \vspace{-0.65cm}
    \caption{\textbf{Training stabilization improves performance and mitigates
    learner degradation.}
    Across environment complexities, WC+HL controls parameter growth while
    maintaining strong learning signals, preserves value--return ordering, and
    mitigates late representation degradation in the more challenging settings.
    These changes accompany consistently higher solve rates and returns.}
    \vspace{-0.25cm}
    \label{fig:stabilized_optimization}
\end{figure*}

\subsection{Effects on Optimization and Curriculum Dynamics}
\label{sec:optimization_dynamics}

We denote WC+HL as the \emph{Stabilized} learner and compare it with baseline
PLR across \textcolor{smallcolor}{\textbf{Small (S)}},
\textcolor{mediumcolor}{\textbf{Medium (M)}}, and
\textcolor{largecolor}{\textbf{Large (L)}} environments. We first examine
whether stabilization mitigates the learner-side degradation identified in
Sec.~\ref{sec:diagnosis}. We then analyze how these changes affect the
curriculum realized by the unchanged PLR rule. As shown in \autoref{fig:stabilized_optimization}, the intervention controls
parameter growth across all complexity levels. The Stabilized learner also
maintains substantial gradient norms while achieving higher solve rates and
returns. This pattern suggests that stabilization does not operate simply by
suppressing gradient magnitudes.
Instead, it allows the learner to remain
responsive to experience while preventing parameter scale from growing
unconstrained. 

\paragraph{Value-prediction dynamics.}
\autoref{fig:stabilized_optimization} further shows that WC+HL does not
stabilize the critic by simply shrinking its outputs. Instead, the
intervention changes how value predictions evolve throughout training.
Prediction and target means follow broadly similar trends, while
prediction dispersion is better preserved than in the baseline,
particularly in \textcolor{largecolor}{Large}. In the baseline,
prediction variance contracts as training progresses, most notably
in the more challenging configurations. However, reduced dispersion
alone does not establish critic degradation, since the scale and
distribution of value targets also differ across environments.
A more informative distinction emerges from the correspondence
between predicted values and empirical returns. Under the baseline,
value--return rank correlation deteriorates substantially in
\textcolor{largecolor}{Large} and, to a lesser extent, in
\textcolor{mediumcolor}{Medium}, indicating that the critic becomes
less effective at preserving the relative ordering of outcomes.
In contrast, WC+HL maintains considerably stronger and more stable
rank correlation across all three configurations, including those
where baseline performance stagnates. 
These results suggest that stabilization does more than control
the scale or variability of value predictions. WC+HL maintains
a stronger rank correlation between predicted values and
empirical returns, indicating that the critic better preserves
the relative ordering of outcomes throughout training.

\paragraph{Representation dynamics.}
As shown in \autoref{fig:stabilized_optimization}, WC+HL also
substantially alters representation dynamics, although its effects
differ across environment complexities. In
\textcolor{largecolor}{Large}, the baseline initially exhibits
high effective rank but undergoes a pronounced decline during
later training, alongside weak performance and deteriorating
value--return correlation. WC+HL mitigates this late-training
decline, maintaining a more stable effective rank throughout
training. A similar distinction emerges in representation
similarity. In \textcolor{largecolor}{Large} and
\textcolor{mediumcolor}{Medium}, the baseline exhibits increasing
feature alignment during later training, whereas WC+HL maintains
lower similarity. These results indicate that stabilization
mitigates changes in representation structure associated with
poor learning in the more challenging configurations.

Interestingly, the relationship between effective rank and
performance differs across configurations. In
\textcolor{smallcolor}{Small} and
\textcolor{mediumcolor}{Medium}, WC+HL achieves higher returns
despite exhibiting lower effective rank than the baseline
toward the end of training. This suggests that higher effective
rank alone does not consistently correspond to better
performance. In contrast, the pronounced decline in effective
rank under the \textcolor{largecolor}{Large} baseline coincides
with deteriorating value ordering and substantially weaker
learning, while WC+HL maintains more stable representations
and stronger performance. Together, these observations suggest
that the evolution of representation structure during training
may be more informative than effective rank considered in
isolation. The representation diagnostics further complement
the value-prediction results, showing that WC+HL maintains
stronger value ordering while mitigating late-training
representation degradation in the more challenging
configurations. Although these changes accompany improved
performance, the results do not establish that preserving
representation structure directly causes the observed gains.

\paragraph{Effects on the realized curriculum.}

\begin{wrapfigure}[17]{r}{0.60\columnwidth}
    \centering
    \vspace{-0.7cm}
    \includegraphics[width=\linewidth]
    {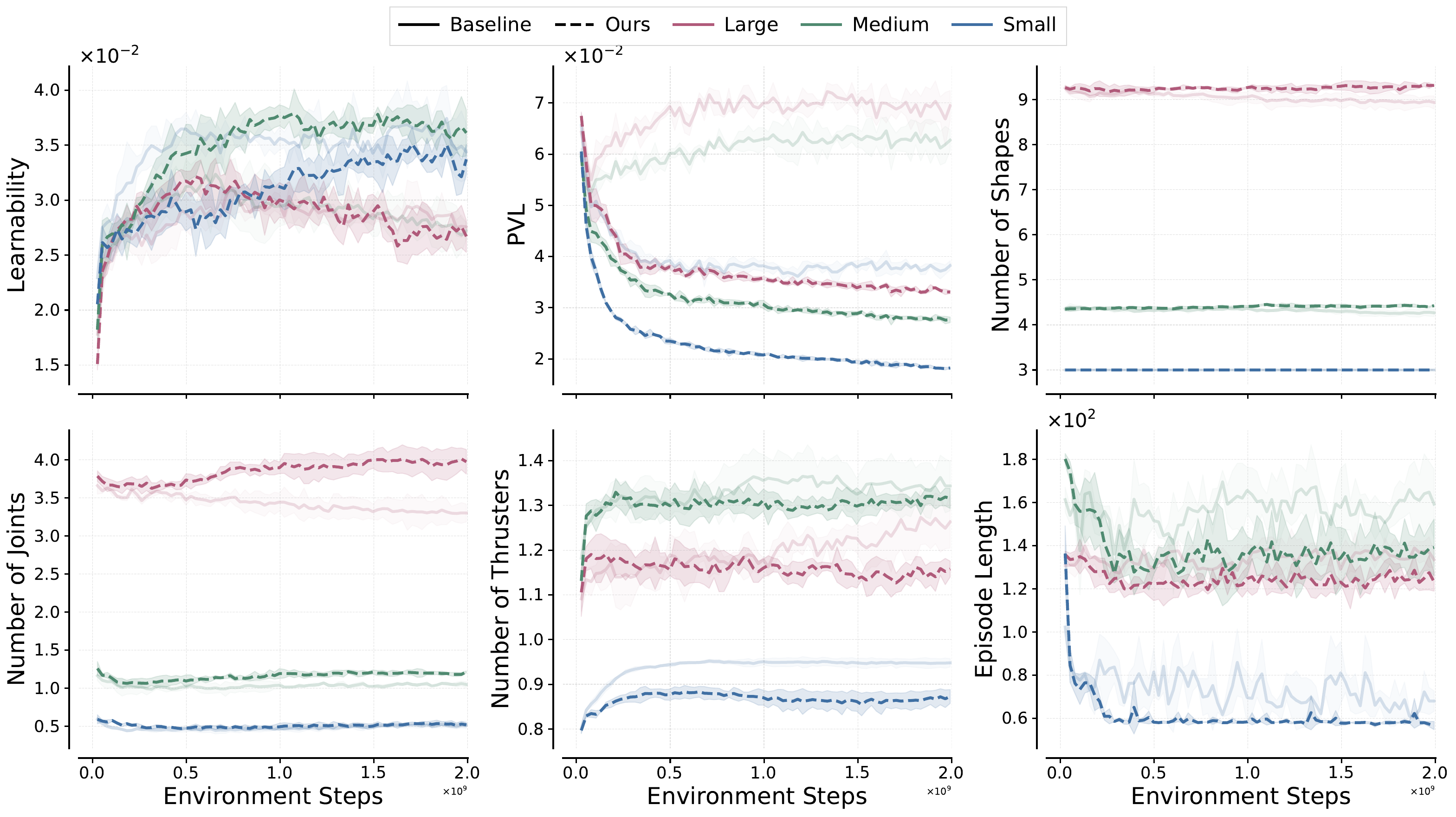}
    \vspace{-0.7cm}
    \caption{\textbf{Training stabilization changes the curriculum realized by
    PLR.}
    Although the replay rule remains fixed, WC+HL changes student-dependent
    curriculum signals and shortens episodes while the measured structural
    properties of replayed levels remain broadly comparable.}
    \label{fig:stabilized_curriculum}
\end{wrapfigure}

The preceding results show that WC+HL improves learner performance and mitigates several optimization and representation pathologies. Under PLR, these changes can also influence curriculum behavior because replay priorities are computed from student-dependent value statistics. As shown in
\autoref{fig:stabilized_curriculum}, WC+HL produces lower PVL scores across
complexity levels, indicating that the same PLR rule receives different
value-based priority signals from a more trainable student. 

Learnability changes non-uniformly across complexity levels, remaining broadly
comparable in \textcolor{smallcolor}{Small}, increasing in Medium, and decreasing late in \textcolor{largecolor}{Large}.
Thus, stabilization does not simply shift the curriculum toward uniformly more
learnable levels. The realized environments also retain broadly comparable
measured structural complexity: shape counts remain similar, joint counts are
comparable or higher under stabilization, and thruster counts change only
moderately. Episode lengths become shorter, particularly in Small and Medium. Alongside the performance improvements reported above, these results suggest that stabilization increases the suitability of generated levels to provide the student with learning signal, while keeping them structurally comparable rather than uniformly simpler.

\begin{tcolorbox}[
    colback=softgreen,
    leftrule=0.5mm,
    top=0.5mm,
    bottom=0.5mm
]
\paragraph{Takeaway.}
Training-stability interventions improve learner trainability across
environment complexities. Under adaptive UED, these learner-side improvements
also reshape the curriculum realized by an unchanged curriculum rule.
\end{tcolorbox}

\section{Scaling across Training Regimes}
\label{sec:scaling}

\begin{wrapfigure}[22]{r}{0.41\columnwidth}
    \centering
    \vspace{-0.5cm}
    \includegraphics[width=\linewidth]
    {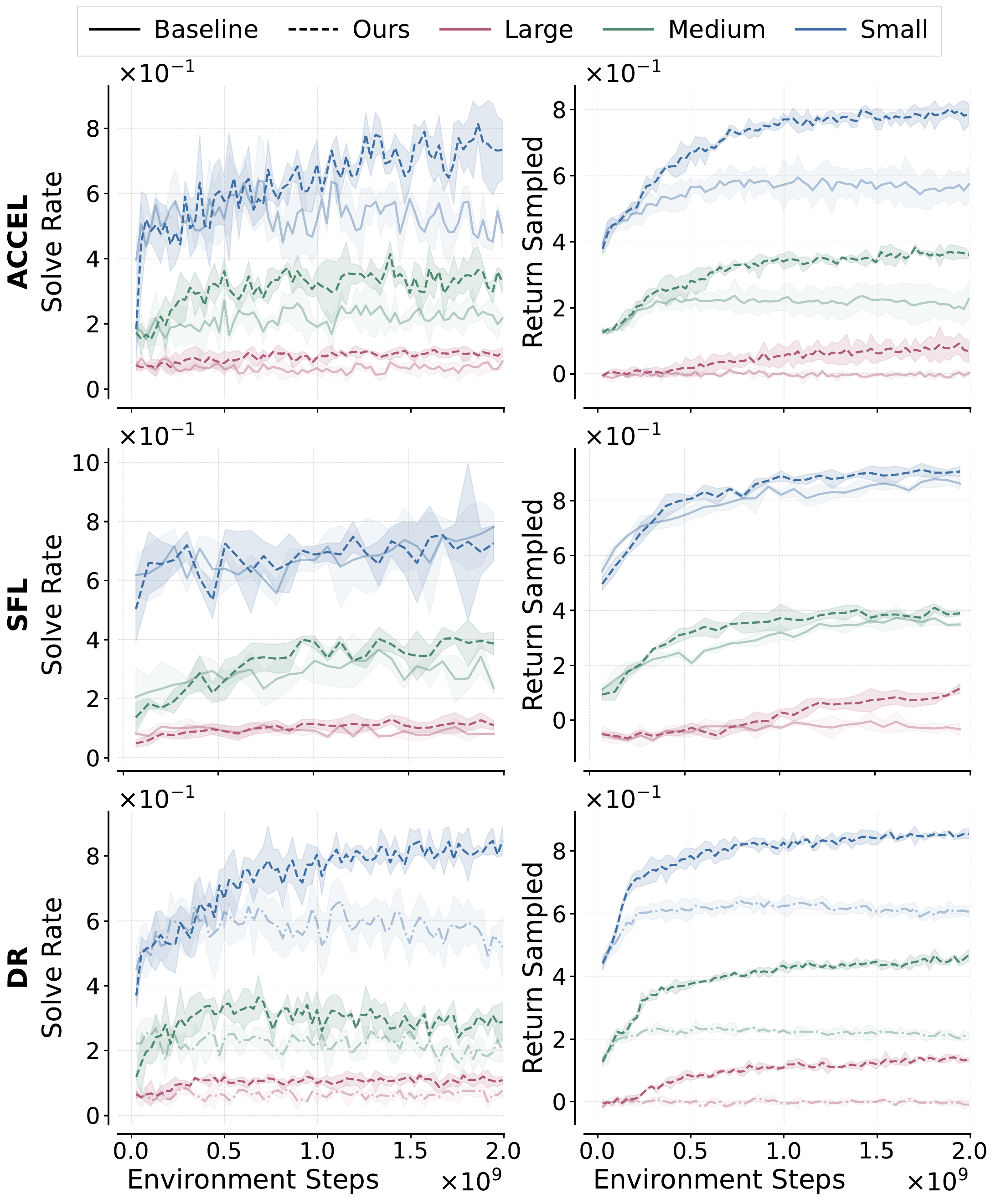}
    \vspace{-0.7cm}
    \caption{\textbf{Stabilization improves performance across
    training-distribution mechanisms.}
    WC+HL improves solve rate and return for DR, SFL, and ACCEL across Kinetix
    complexity levels.}
    \label{fig:across_agents}
    %\vspace{-10cm}
\end{wrapfigure}

The preceding experiments show that controlling parameter scale and value-loss
sensitivity improves PLR training. We now test whether these benefits persist
across three dimensions: the mechanism constructing the training distribution,
the scale of training, and the evaluation distribution. Unless stated
otherwise, \emph{Stabilized} denotes WC+HL, using the configuration selected in
Sec.~\ref{sec:stabilization}. Within each comparison, the baseline and
Stabilized learners use the same curriculum rule, architecture, and
environment-interaction budget.

\paragraph{Across training-distribution mechanisms.}
We extend the evaluation beyond PLR to the domain-randomization (DR) baseline
\citep{jakobi1997evolutionary,tobin2017domain}, SFL
\citep{rutherford2024no}, and ACCEL
\citep{parkerholder2022evolving}. These approaches construct their training
distributions differently: DR samples directly from the generator, SFL selects
levels according to estimated learnability, and ACCEL expands the task
distribution through environment editing. As shown in
\autoref{fig:across_agents}, stabilization yields consistent gains across these
mechanisms and Kinetix complexity levels. The effect is therefore not specific
to PLR's replay rule, although its magnitude varies across curriculum
mechanisms.

\begin{figure*}[!h]
    \centering
    \includegraphics[width=\textwidth]
    {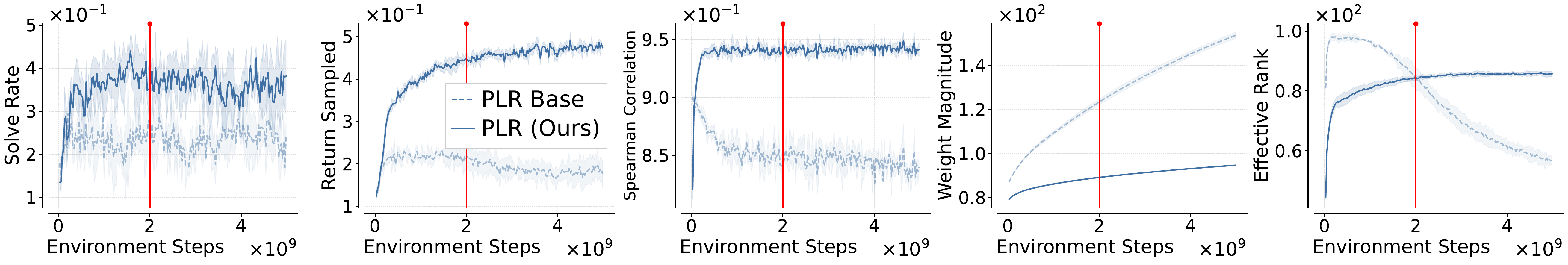}
    \vspace{-0.6cm}
    \caption{\textbf{Stabilization extends the useful training horizon of PLR.}
    Beyond the original two-billion-step budget, marked in orange, baseline
    return and learner diagnostics deteriorate. WC+HL maintains solve rate,
    continues improving return, and preserves value ordering and effective
    rank through five billion interactions.}
    \vspace{-0.25cm}
    \label{fig:long_horizon2}
\end{figure*}

\paragraph{Scaling the training horizon.}
Additional interaction is useful only while the learner can convert experience
into improved behavior. We therefore extend PLR training on Small environments
from two to five billion interactions. As shown in
\autoref{fig:long_horizon2}, the baseline obtains little benefit beyond the
original budget: return declines, parameter magnitude continues to grow,
value--outcome correspondence remains weak, and effective rank deteriorates.
The Stabilized learner instead maintains solve rate, continues to gain return,
and preserves value ordering and representation rank. The performance gap
therefore widens beyond two billion interactions, showing that conclusions
about a UED method can depend on the training horizon at which it is evaluated.

\begin{figure*}[!t]
    \centering
    \includegraphics[width=\textwidth]
    {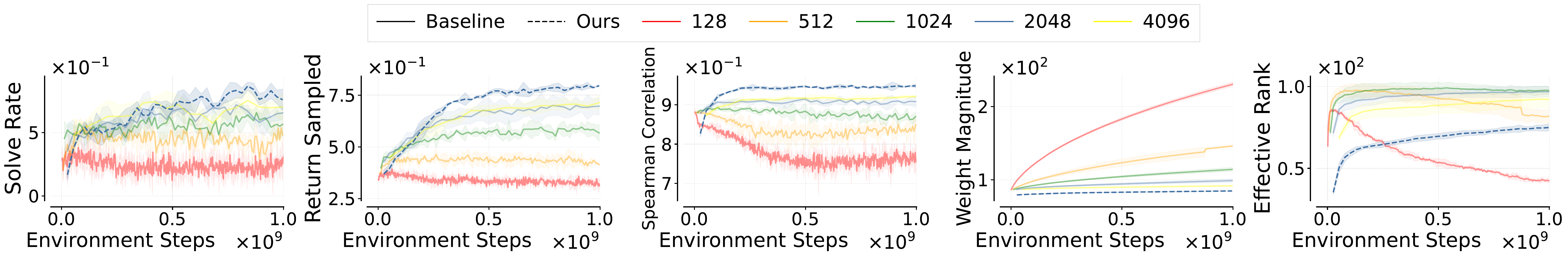}
    \includegraphics[width=\textwidth]
    {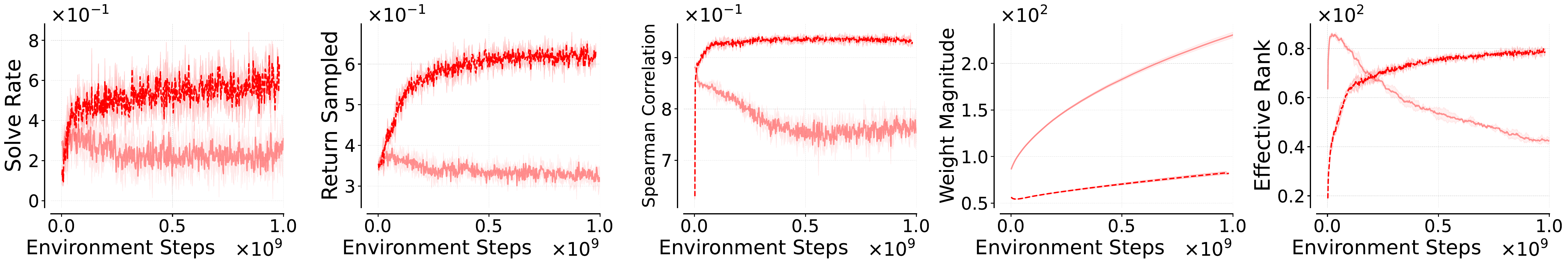}
    \vspace{-0.6cm}
    \caption{\textbf{Stabilization improves PLR across parallelism regimes.}
    \textbf{Top:} Increasing parallelism improves performance and learner
    diagnostics up to the default 2048-environment setting, with little additional
    gain at 4096 environments. At the default scale, WC+HL achieves stronger
    performance together with healthier training dynamics than the corresponding
    baseline.
    \textbf{Bottom:} Under limited parallelism, with only 128 environments, WC+HL
    still improves performance, preserves value ordering and effective rank, and
    controls parameter growth.}
    \vspace{-0.25cm}
    \label{fig:parallelism_scaling2}
\end{figure*}

\paragraph{Stabilization across parallelism regimes.}
Recent work shows that increasing PPO rollout batch size can prevent a
particular form of learning stagnation in open-ended Kinetix training
\citep{beukman2026preventinglearningstagnationppo}. We study how this scaling
strategy interacts with the learner-side degradation considered here by varying
the number of parallel environments used by PLR on Small while holding total
interaction fixed. As shown in \autoref{fig:parallelism_scaling2}, increasing
parallelism yields better performance, value ordering, and effective rank up to
the default 2048-environment setting, with little additional gain at 4096
environments. At the default scale, WC+HL further strengthens performance while
maintaining healthier training dynamics than the corresponding baseline.
Moreover, with only 128 environments, stabilization substantially raises solve
rate and return while controlling parameter growth and representation
degradation. These results suggest that rollout scaling and learner
stabilization are complementary. Increasing parallelism can enhance PPO
updates, but its benefits eventually saturate, while stabilization remains
effective both at the default scale and when large-scale parallelism is
unavailable.

\begin{wrapfigure}[17]{r}{0.46\columnwidth}
    \centering
    \vspace{-0.5cm}
    \includegraphics[width=\linewidth]
        {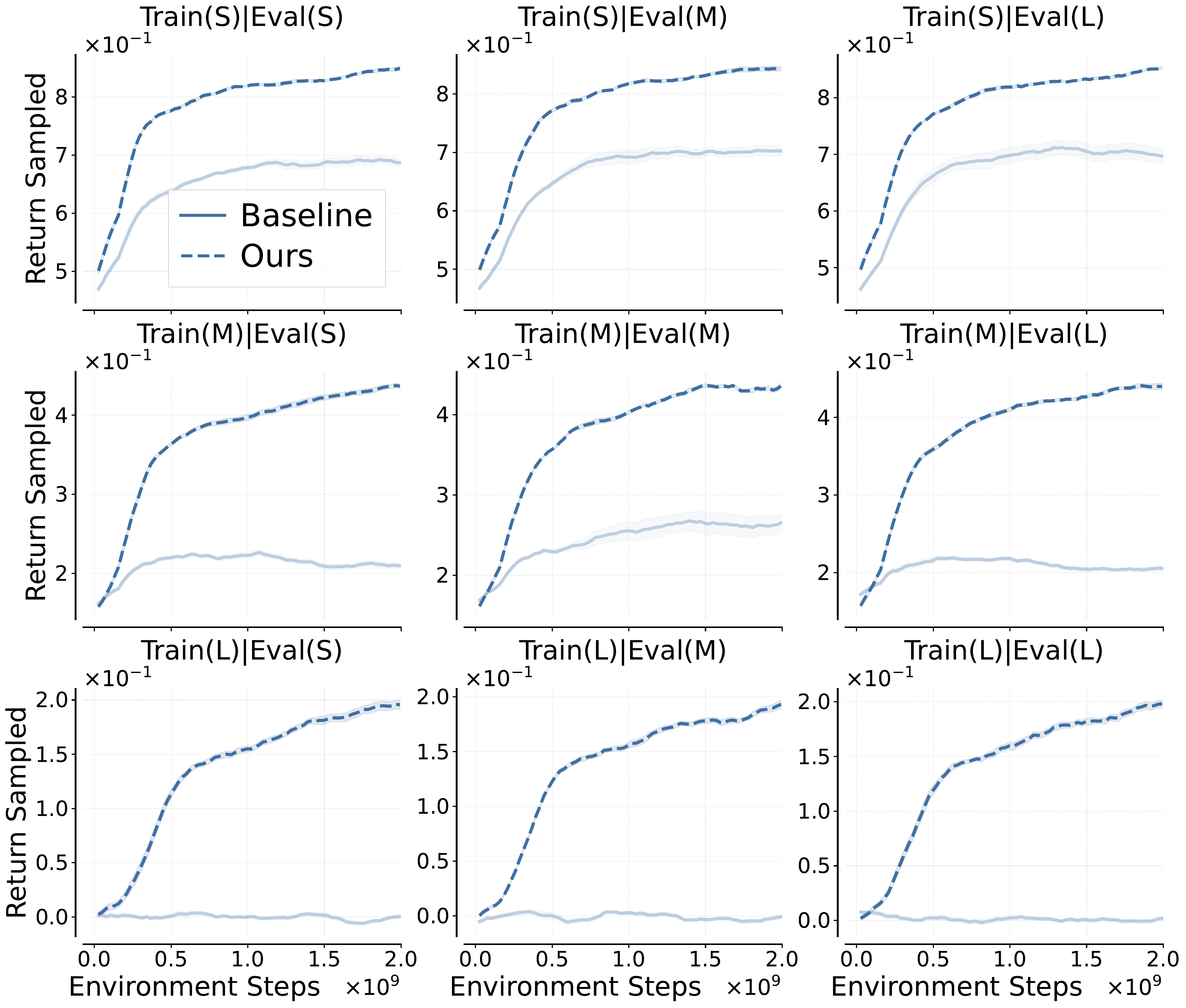}
        \vspace{-0.6cm}
        \captionof{figure}{
        \textbf{Generalization to randomly generated tasks.} Zero-shot sampled return across training and evaluation complexities.
        }
    \label{fig:sampled_cross_complexity1}
     %\vspace{1cm}
\end{wrapfigure}

\paragraph{Cross-complexity generalization.}
We next test whether stabilization improves only the training regime or also
transfers across task complexity. Following the Kinetix protocol
\citep{matthewskinetix}, agents trained on \textcolor{smallcolor}{Small}, \textcolor{mediumcolor}{Medium}, or \textcolor{largecolor}{Large} environments
are evaluated without adaptation on all three complexity levels. \autoref{fig:sampled_cross_complexity1} reports episodic return on generated
evaluation levels, while \autoref{fig:heldout_cross_complexity} reports solve
rate on human-designed holdout tasks. Stabilization improves performance at matched complexity levels and transfer across different complexity levels. The largest gains occur for agents trained on Medium and
Large, where the corresponding baselines exhibit the strongest optimization
and representation degradation. The consistent improvements in solve rate and
sampled return show that stabilization benefits both successful task completion
and graded progress.

\paragraph{Robustness to sequential distribution shifts.}
Finally, we evaluate robustness to changes in the training distribution over time.
Following the Kinetix lifelong-learning protocol
\citep{matthewskinetix}, PLR is trained sequentially on
\textcolor{smallcolor}{\textbf{Small}},
\textcolor{mediumcolor}{\textbf{Medium}}, and
\textcolor{smallcolor}{\textbf{Small}} environments. The optimizer state and
PLR buffer are retained across transitions for both learners, and performance
is continuously measured on the Medium human-designed holdout set.

\begin{wrapfigure}[10]{r}{0.5\columnwidth}
    \centering
    \vspace{-1cm}
    \includegraphics[width=\linewidth]
    {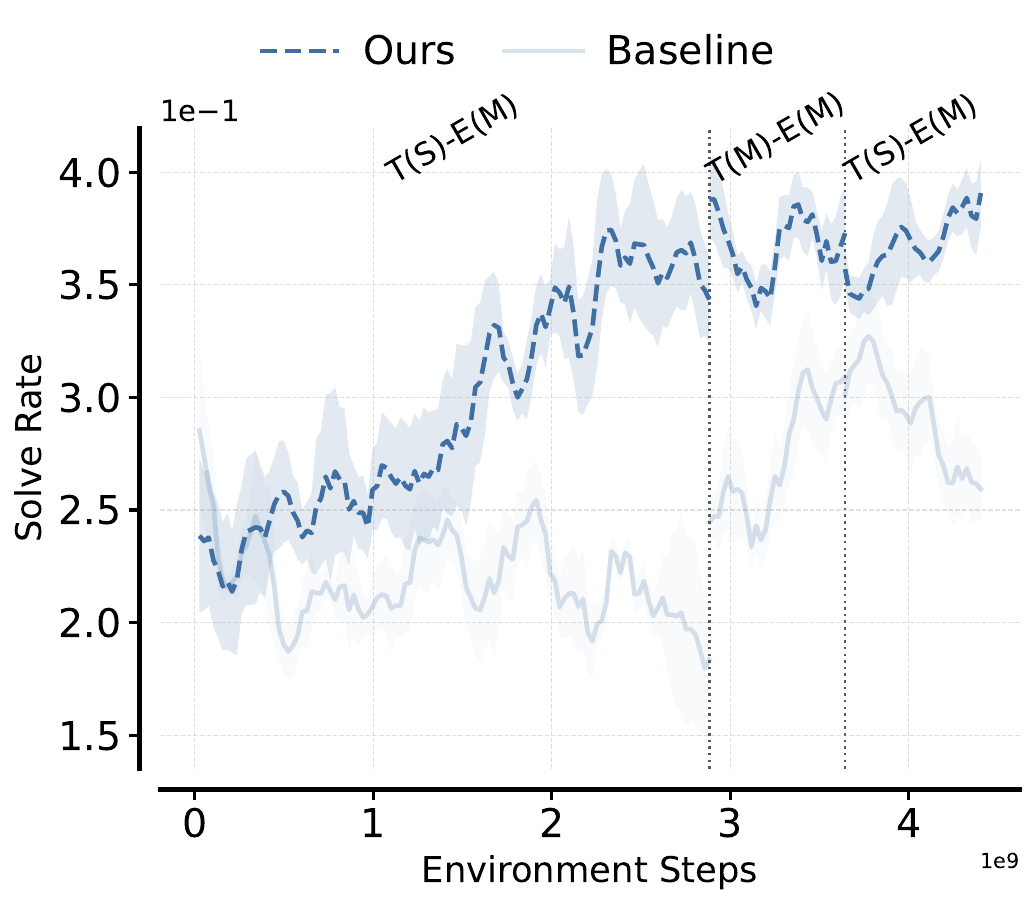}
    \vspace{-0.7cm}
    \caption{\textbf{Stabilization improves retention under sequential
    distribution shifts.} \emph{Ours} transfers more strongly to the Medium holdout set and
    better preserves that performance after training returns to Small. Dotted
    lines mark distribution changes.}
    \label{fig:distribution_shifts}
    \vspace{-2cm}
\end{wrapfigure}

As shown in \autoref{fig:distribution_shifts}, the Stabilized learner transfers
more strongly to Medium tasks during the initial Small stage and maintains its
advantage after training switches to Medium. When training returns to Small,
the baseline loses much of the performance gained during the Medium stage,
whereas the Stabilized learner retains more of that improvement. Because both
learners use the same replay and state-retention protocol, this difference
cannot be attributed to additional rehearsal or an explicit continual-learning
objective. The result therefore supports improved retention on the Medium
holdout set under sequential distribution shift.

This pattern is consistent with reduced forgetting, interference, or loss of
plasticity, but the Medium-only evaluation does not distinguish among these
mechanisms or fully characterize the adaptation--retention trade-off. We
therefore interpret the experiment as evidence of improved retention rather
than a complete continual-learning characterization.

\vspace{0.1cm}

\begin{tcolorbox}[
    colback=softgreen,
    leftrule=0.5mm,
    top=0.5mm,
    bottom=0.5mm
]
\paragraph{Takeaway.}
Learner stabilization generalizes beyond PLR, remains effective over longer
training horizons and across parallelism regimes, and improves transfer and
retention across distributions. Learner trainability therefore shapes how
effectively UED systems convert additional scale and task diversity into
sustained performance.
\end{tcolorbox}

\section{Discussion and Conclusion}
\label{sec:discussion}

A better curriculum cannot indefinitely compensate for a learner that is
losing its ability to learn. Our detailed analysis of PLR shows that extended
training can be accompanied by parameter growth, degraded value predictions,
and deteriorating representations. Controlled interventions targeting these
pathologies improve learning, with benefits extending across SFL, ACCEL, and
domain randomization. Weight clipping controls parameter scale, while HL-Gauss
reduces the sensitivity of value learning to scalar target errors. Their
combination yields stronger performance and more stable training dynamics
without changing the curriculum rule or interaction budget.

These results suggest that UED should be viewed not only as a problem of constructing better curricula but also as a coupled system in which the learner determines how useful a curriculum can be. Because adaptive curricula
depend on student-generated signals, learner degradation can affect both how
experience is used and which experience is encountered next. Conversely,
stabilizing the learner changes the training distribution realized by an
otherwise unchanged curriculum rule. Apparent limitations of curriculum design
may therefore partly reflect degradation in the learner used to evaluate it.
Training dynamics should thus be considered part of curriculum design, rather
than treating the learner update as a fixed component.

This perspective also has implications for evaluation. Performance at a fixed
training horizon may hide whether a learner remains capable of benefiting from
additional experience. Our long-horizon and parallelism results show that more
interaction or rollout scale does not automatically translate into continued
learning. Monitoring learner-side dynamics alongside return can therefore help
distinguish limits of the curriculum from limitations of the learner. Scaling the
curriculum and preserving learner trainability should not be treated as
independent problems.

\textbf{Limitations.}
Our evidence is limited to Kinetix and PPO, Moreover, hyperparameter sensitivity,
probe choices, and intervention interactions constrain causal attribution.
We do not claim that weight clipping and HL-Gauss are uniquely effective.
Continual learning is a particularly challenging extension for UED, where
curriculum-induced non-stationarity compounds explicit task shifts. Broader
validation and adaptive stabilization remain important directions for future
work.

 \section*{Acknowledgments}
The authors would like to thank Gopeshh Subbaraj for valuable discussions during the preparation of this work. Clare Lyle deserves a special mention for providing valuable feedback on an early draft of the paper. The research was enabled in part by computational resources provided by the Digital Research
Alliance of Canada (\url{https://alliancecan.ca}) and Mila (\url{https://mila.quebec}). Aaron Courville acknowledges funding from the Institute of Information \& Communications Technology Planning \& Evaluation (IITP) through a grant funded by the Korean government (MSIT) (No. RS-2024-00457882, National AI Research Lab Project). Pablo Samuel Castro acknowledges funding from NSERC Discovery Grant. We want to acknowledge funding support from Google and CIFAR AI. We would also like to thank the Python community \citep{van1995python, 4160250} for developing tools that enabled this work, including NumPy \citep{harris2020array}, Matplotlib \citep{hunter2007matplotlib}, Jupyter \citep{2016ppap}, and Pandas \citep{McKinney2013Python}.

\section*{AI use statement}

In this work, we used generative AI tools to assist with paragraph-level writing and polishing, discuss and refine the structure and flow of the introduction, identify related work, and write and edit code. We also used generative AI tools to assist with early drafts of some sections, which the authors later substantially revised or replaced. We did not use generative AI tools for research ideation. Other disclosure categories, such as generating synthetic datasets or proving mathematical claims, are not applicable to this work
We have reviewed the AI-assisted work. We reviewed LLM-generated code and read and incorporated suggested related work papers manually.
We take responsibility for the final content of this work, including text, claims or artifacts produced with the aid of generative AI.

\bibliography{iclr2027_conference}
\bibliographystyle{iclr2027_conference}

% % %%%%%%%%%%%%%%%%%%%%%%%%%%%%%%%%%%%%%%%%%%%%%%%%%%%%%%%%%%%%

\clearpage
\newpage

\appendix
\onecolumn

\etocdepthtag.toc{appendix}
\etocsettocstyle{

\section*{Appendix Contents}}{} % Title of the TOC
\setcounter{tocdepth}{3}
\etocsettagdepth{mtoc}{none}           % Hide main text sections
\etocsettagdepth{appendix}{subsection}  % Show appendix sections (adjust depth as needed)
\tableofcontents                        % Print the TOC

\clearpage
\newpage
\input{appendix}

\end{document}

%% file: appendix.tex
\appendix

\section{Related Work}
\label{sec:relatedwork2}

UED methods construct training distributions by generating, selecting, or
replaying environments using signals derived from the student. PAIRED frames
environment generation as an adversarial teacher--student game
\citep{dennis2020emergent}, PLR and replay-guided environment design prioritize
previously encountered levels using estimates of regret or learning potential
\citep{jiang2021prioritized,jiang2021replay}, ACCEL expands the curriculum
through environment editing \citep{parkerholder2022evolving}, and SFL selects
environments estimated to be learnable by the current student
\citep{rutherford2024no}. Kinetix evaluates these mechanisms in a large
and heterogeneous space of physics-based tasks, demonstrating the importance
of curriculum choice while also revealing substantial degradation with
increasing environment complexity \citep{matthewskinetix}. Related work has
improved the stability of particular teacher--student algorithms
\citep{mediratta23a} or studied UED objectives and convergence from
an optimization perspective \citep{monette2025an}. These works
primarily address how curricula are constructed or optimized. Separately, deep
RL research has documented parameter growth, feature-rank deterioration,
dormant neurons, representation collapse, and loss of plasticity during
prolonged training
\citep{kumar2021implicit,nikishin2022primacy,sokar2023dormant,
lyle2023und}. In PPO, representation degradation can undermine
effective trust-region behavior \citep{moalla2024no}, while on-policy data
collection choices can substantially alter network optimization and plasticity
\citep{mayor2025impact}. Existing interventions can mitigate particular
failure modes, including parameter growth \citep{elsayed2024weight} and
sensitivity to scalar value targets \citep{farebrother2024stop}, but these
phenomena have largely been studied outside UED. Our work connects these
literatures by examining how the student's optimization dynamics evolve under
curriculum--student coupling and whether addressing the resulting pathologies
improves existing UED methods without redesigning their curricula.

\section{Learner Trainability as a Design Principle for UED}
\label{sec:design_principle}

UED has largely been formulated as a problem of constructing effective training
distributions. PAIRED generates environments through an adversarial
teacher--student game \citep{dennis2020emergent}, PLR and replay-guided methods
prioritize previously encountered levels using estimates of learning potential
or regret \citep{jiang2021prioritized,jiang2021replay}, ACCEL expands the
curriculum through environment editing
\citep{parkerholder2022evolving}, and SFL selects environments estimated to be
learnable by the current student \citep{rutherford2024no}. These methods
differ substantially in how they construct curricula, but they are commonly
evaluated through the performance ultimately achieved by a shared class of
student learners. Related work has improved the stability of particular
teacher--student algorithms \citep{mediratta23a} and studied UED objectives and
convergence from an optimization perspective \citep{monette2025an}. Our results
highlight a complementary concern. Even when a curriculum identifies useful
training environments, degradation within the student can prevent that
experience from producing further capabilities.

This distinction is important because the observed performance of a UED method
reflects two coupled processes. The curriculum determines which environments
generate experience, while the student determines how effectively that
experience is converted into learning. Deep RL research outside UED has shown
that prolonged optimization can produce parameter growth, feature-rank
deterioration, dormant neurons, representation collapse, and loss of plasticity
\citep{kumar2021implicit,nikishin2022primacy,sokar2023dormant,
lyle2023und,lyle2024normalization}. These failures also arise in PPO, where representation degradation
can weaken effective trust-region behavior \citep{moalla2024no}, and can depend
strongly on how on-policy experience is collected and reused
\citep{mayor2025impact}. Under UED, such degradation has a further consequence.
Changes in the student can alter the environments subsequently generated,
selected, or replayed, allowing learner optimization to influence its own
future training distribution. Apparent limitations of a curriculum mechanism
may therefore partly reflect limitations of the student through which it is
evaluated.

Learner trainability should consequently be treated as a design and evaluation
principle for UED. Comparisons between curriculum mechanisms should report not
only final return and properties of the generated environments, but also
whether their students remain capable of learning as task complexity and
interaction budgets increase. This motivates tracking parameter and update
statistics, value predictions, neuronal activity, and representation quality
alongside performance. Evaluations should also distinguish the curriculum rule
from the curriculum it ultimately realizes, since two students trained under
the same adaptive rule can induce different environment distributions as their
behaviors diverge. Developing stronger curriculum mechanisms remains essential,
but their benefits may be obscured when fundamental optimization failures
dominate training. Revisiting existing UED methods with more stable learners
may therefore reveal capabilities that were previously hidden, while future
methods could account not only for which environments appear informative, but
also for whether the student remains sufficiently trainable to benefit from
them.

\section{Selected interventions.}
\label{app:selected}

Prior work has identified several optimization and representation
pathologies in deep RL, including parameter growth, dormant neurons,
representation collapse, and limited effective use of network
capacity \citep{nikishin2022primacy,sokar2023dormant,
mayor2025impact,obando2026simplicial,
pmlr-v235-obando-ceron24a,
pmlr-v235-obando-ceron24b,lee25u,
pasand2026stable,liu2026measure}.
These findings have motivated interventions that target different
aspects of learner trainability, including parameter regularization,
representation learning, and alternative value-learning objectives
\citep{elsayed2024weight,castanyer2025stable,farebrother2024stop,liu2025neuroplastic,liu2025the,pasand2026stable}.
Here, we focus on two complementary interventions: weight clipping,
which directly controls critic parameter scale, and HL-Gauss,
which changes the value-learning objective.

Weight clipping constrains the critic parameters $\phi$ after every optimizer
update~\citep{elsayed2024weight}. Given unconstrained parameters
$\widetilde{\phi}_{k+1}$, we apply
$\phi_{k+1}=\Pi_{\mathcal{C}_c}(\widetilde{\phi}_{k+1})$, where
$\mathcal{C}_c=\{\phi:\|\phi\|_\infty\leq c\}$. This directly controls
parameter scale but does not bound gradients or function changes. HL-Gauss replaces scalar squared-error regression with a histogram
cross-entropy loss~\citep{farebrother2024stop}. The critic predicts probabilities
$p_\phi(\cdot\mid s)=\operatorname{softmax}(h_\phi(s))$ over a fixed set of
ordered bins. A scalar target $y$ is converted into a categorical label
$m^{\mathrm{HL}}(y)$ by integrating a Gaussian centered at $y$ over those
bins, and the scalar value prediction is recovered as the expectation of the
predicted histogram. Unlike distributional RL, HL-Gauss represents a scalar
expected-value target rather than the full return distribution. Its logit
gradient satisfies
$\partial\mathcal{L}_{\mathrm{HL}}/\partial h_j
=p_\phi(z_j\mid s)-m^{\mathrm{HL}}_j(y)\in[-1,1]$, removing the direct linear
dependence of the prediction-head gradient on the raw scalar error. The full
parameter gradient remains dependent on the network Jacobian. Parameter scope,
histogram support, smoothing scale, and sensitivity analyses are provided in
Apps.~\ref{app:weight_clipping} and~\ref{app:categorical_value}.

\section{Interventions}
\label{app:interventions}

\subsection{Weight Clipping}
\label{app:weight_clipping}
We implemented the weight clipping (WC) intervention by varying the scaling factor of the clipping range, using values of $1$, $2$, $4$, and $8$. We compared each configuration against the PLR baseline to evaluate the effect of the clipping range on performance ~\citep{elsayed2024weight}.

\begin{figure*}[!h]
    \centering
    \includegraphics[width=\textwidth]
    {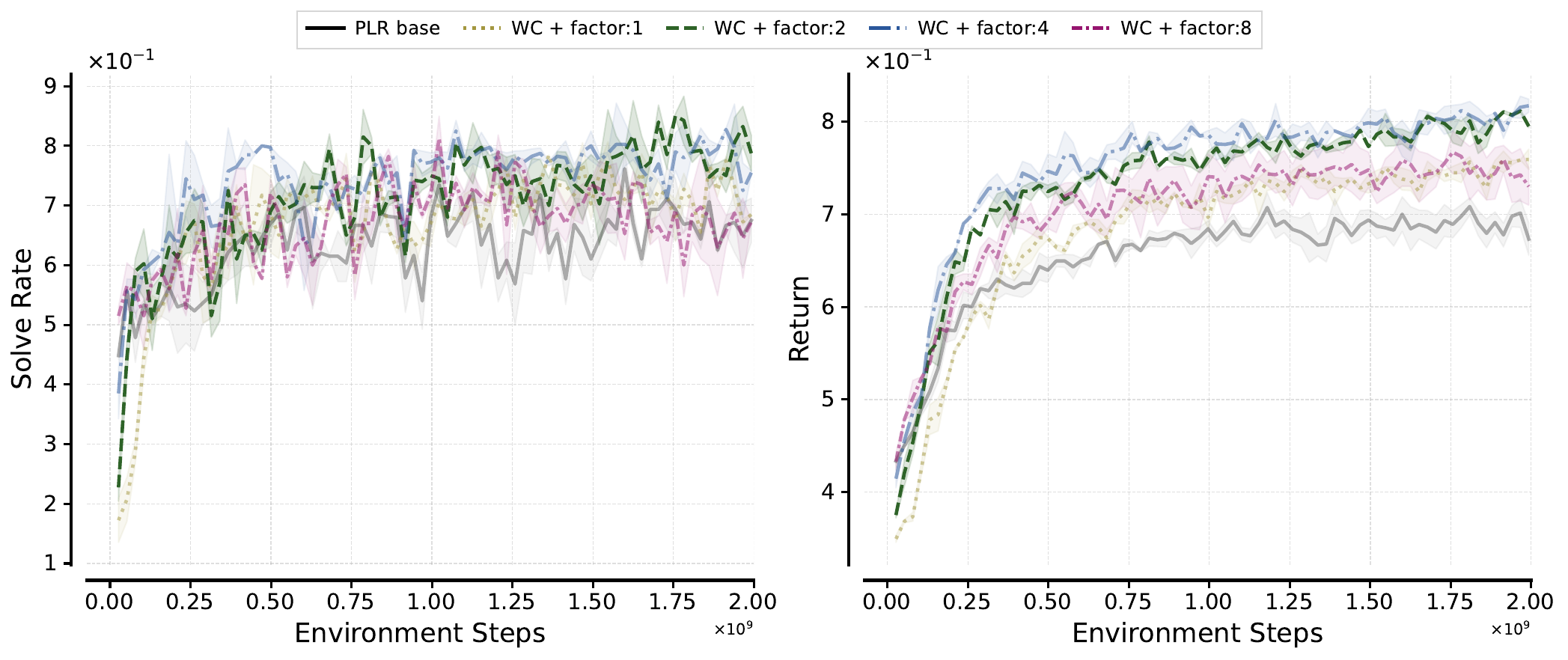}
    \vspace{-0.6cm}
    \caption{\textbf{Solve rate (left) and sampled return (right) over 5 seeds. }We compare the WC intervention on PLR using different scaling factors for the clipping range. Scaling factors below $2$ result in lower performance, while factors of $2$ and $4$ achieve similar performance. Increasing the scaling factor to $8$ leads to a substantial performance degradation.}
    \vspace{-0.25cm}
    \label{fig:wc_test_int}
\end{figure*}

\begin{figure*}[!h]
    \centering
    \includegraphics[width=\textwidth]
    {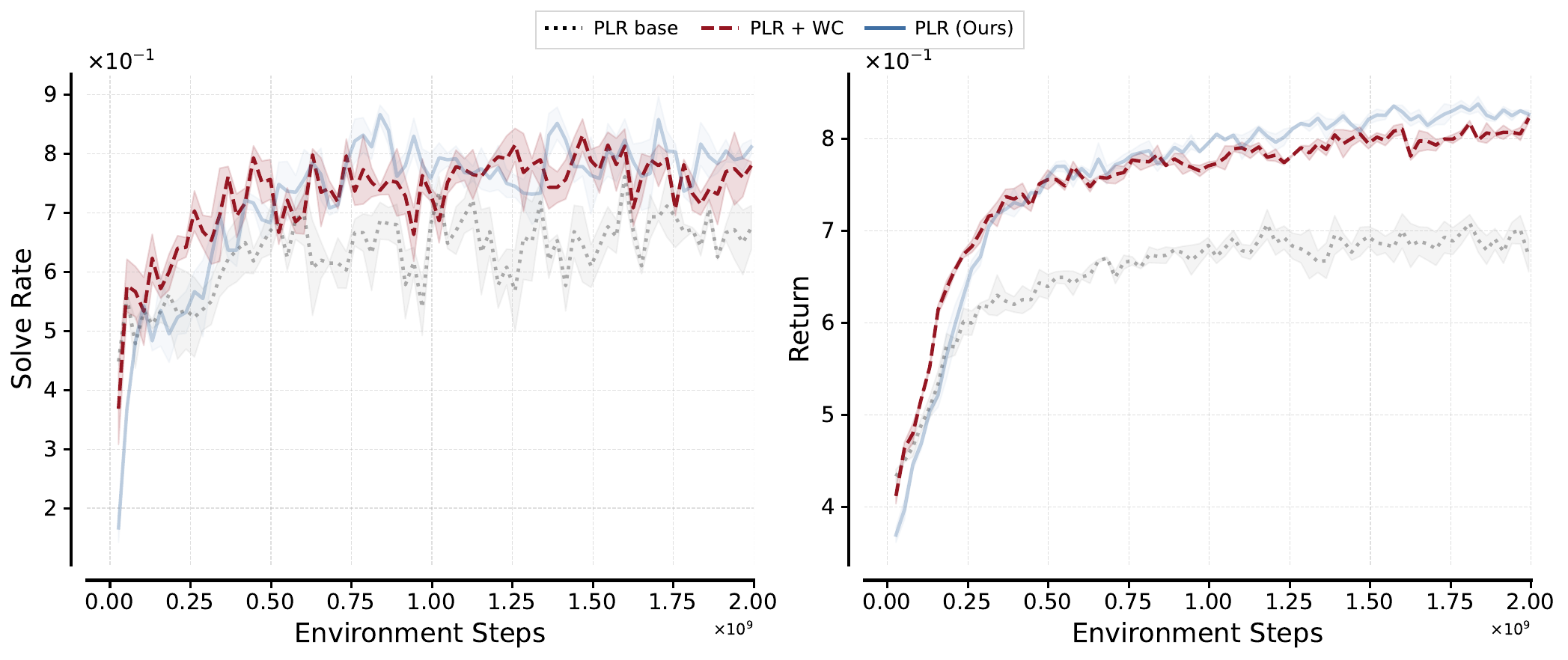}
    \vspace{-0.6cm}
    \caption{\textbf{Solve rate (left) and sampled return (right) over 5 seeds. }We compare the PLR baseline, the best WC configuration using a scaling factor of $2$, and our selected interventions. Our interventions show a modest performance improvement over WC, while both substantially outperform the PLR baseline.}
    \vspace{-0.25cm}
    \label{fig:wc_int}
\end{figure*}

\subsection{Weight Decay}
\label{app:weight_decay}

Based on the results reported in~\cite{castanyer2025stable}, we set the weight decay coefficient to $1\times10^{-2}$. We applied Weight Decay uniformly across the network to regularize its parameters and promote training stability ~\citep{saether2026does}.

\begin{figure*}[!h]
    \centering
    \includegraphics[width=\textwidth]
    {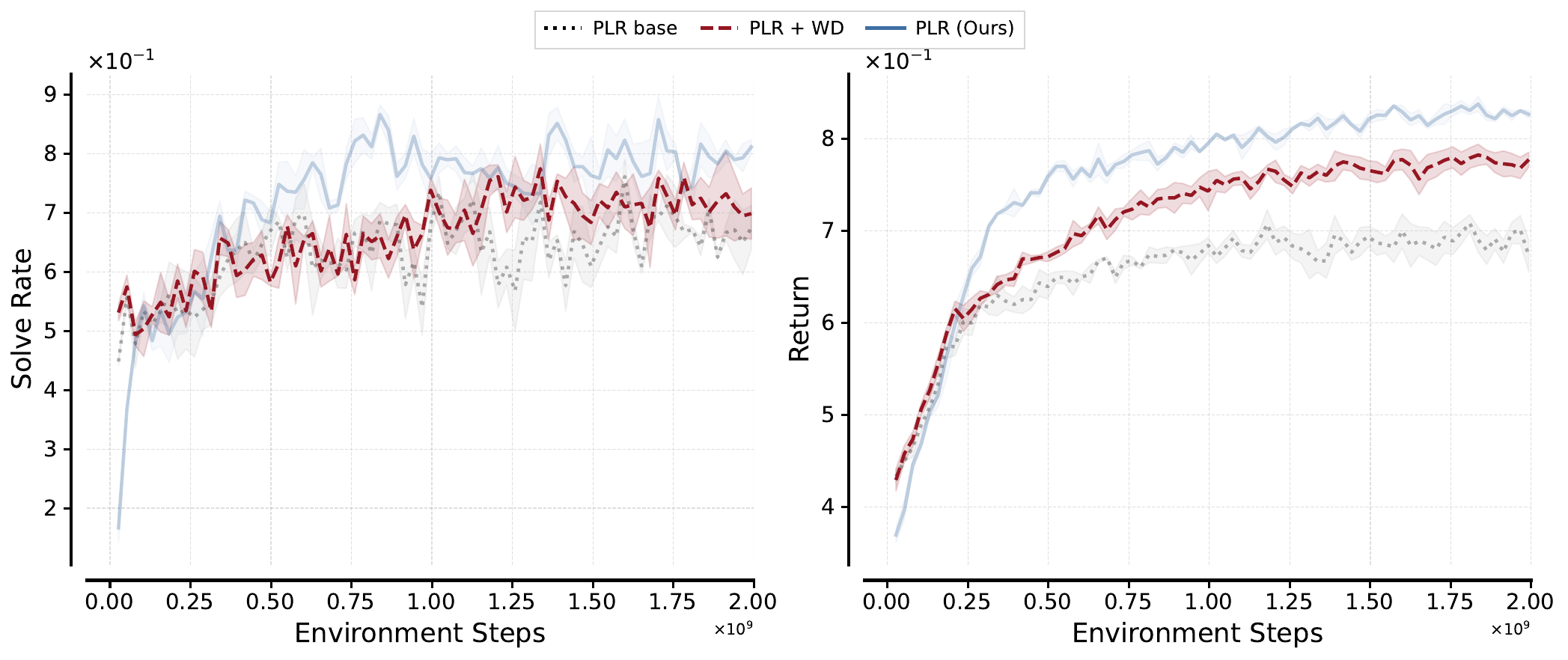}
    \vspace{-0.6cm}
    \caption{\textbf{Solve rate (left) and sampled return (right) over 5 seeds. }We compare the PLR baseline, the WD intervention, and our selected intervention. WD outperforms the PLR baseline, while our intervention achieves higher performance than both WD and the baseline.}
    \vspace{-0.25cm}
    \label{fig:wd_int}
\end{figure*}

\subsection{Spectral Norm}
\label{app:spectral_norm}

We implemented Spectral Normalization to constrain layer norms and stabilize training, evaluating its application to the actor, critic, and both networks. We then compared the best-performing configuration against the PLR baseline and our proposed intervention, allowing us to directly assess their behavior and performance ~\citep{gogianu2021spectral}.

\begin{figure*}[!h]
    \centering
    \includegraphics[width=\textwidth]
    {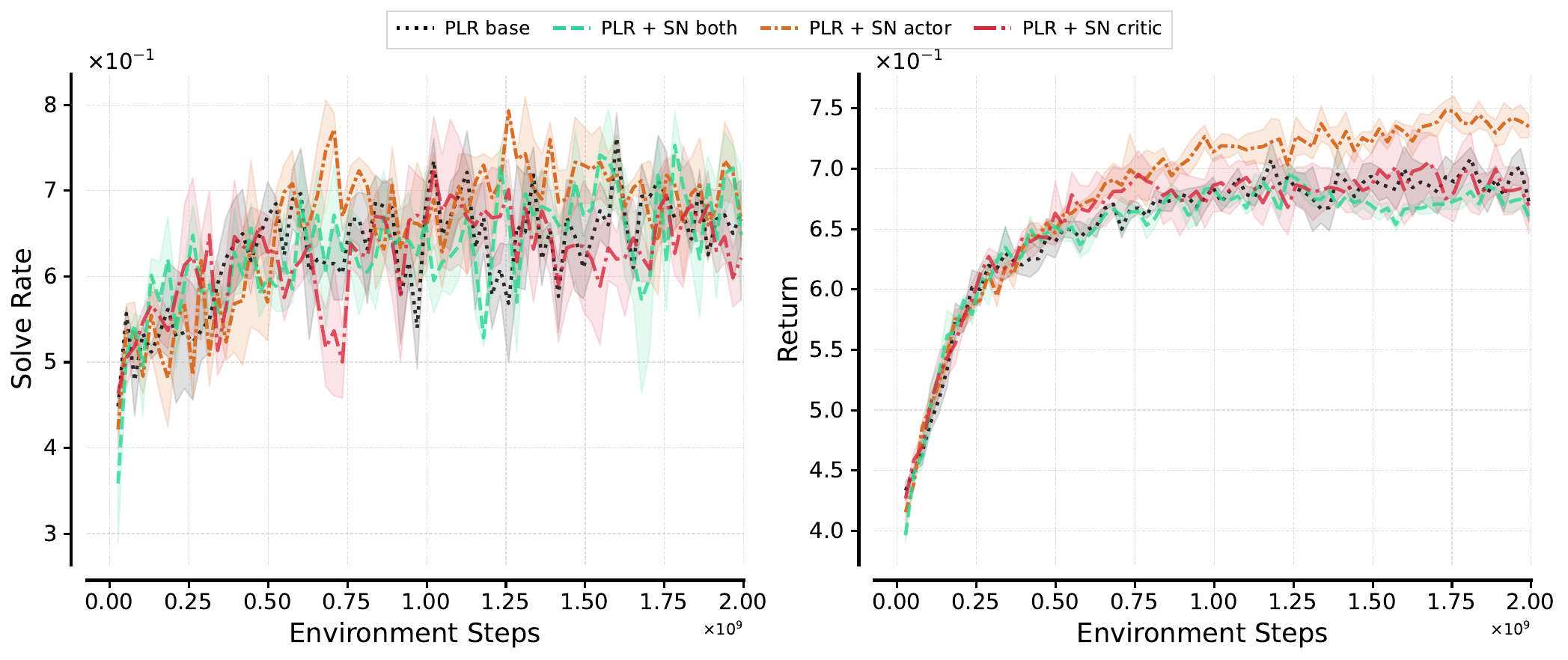}
    \vspace{-0.6cm}
    \caption{\textbf{Solve rate (left) and sampled return (right) over 5 seeds. }We evaluate Spectral Normalization (SN) on PLR by applying it to the actor, the critic, or both networks. Applying SN to the critic or both networks yields similar results, while applying it to the actor achieves higher performance than either configuration.}
    \vspace{-0.25cm}
    \label{fig:sn_test_int}
\end{figure*}

\begin{figure*}[!h]
    \centering
    \includegraphics[width=\textwidth]
    {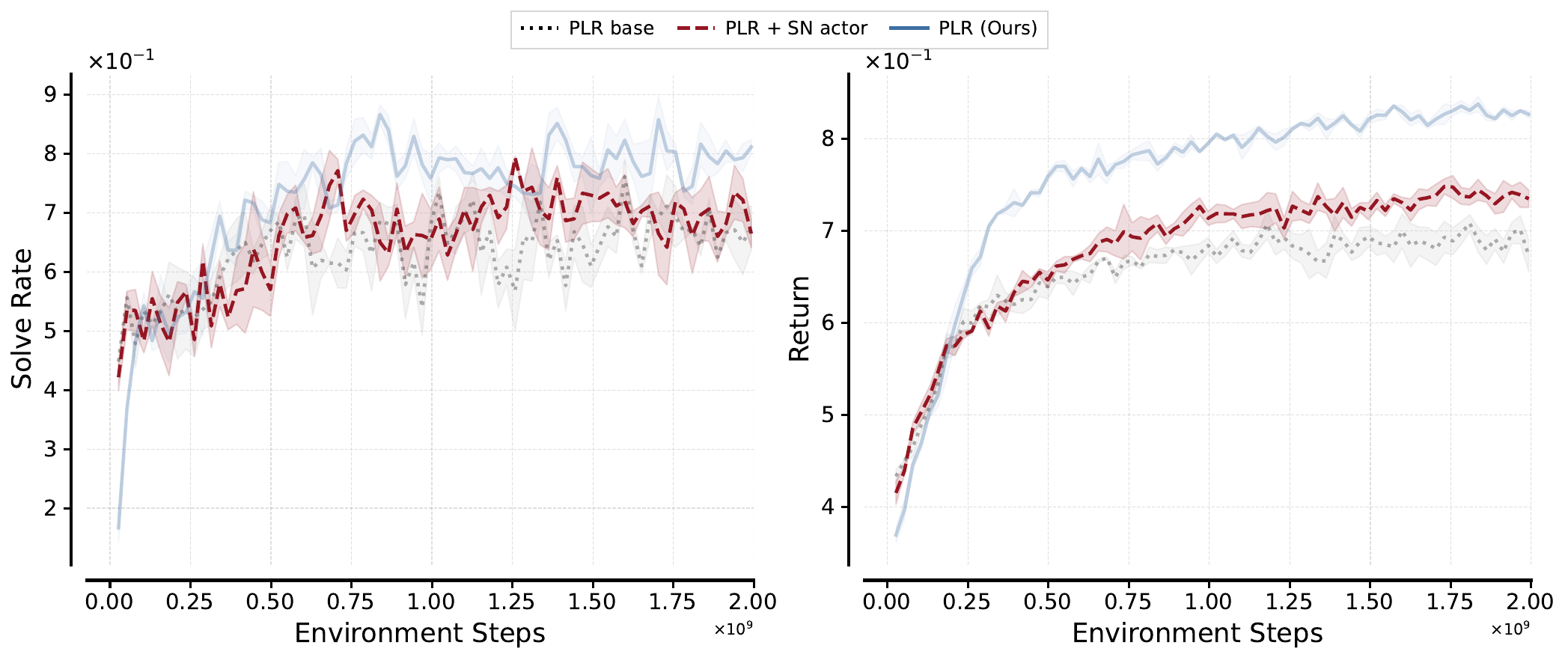}
    \vspace{-0.6cm}
    \caption{\textbf{Solve rate (left) and sampled return (right) over 5 seeds. }We contrast the PLR baseline, the best SN configuration applied to the actor, and our selected intervention. SN on the actor outperforms the PLR baseline, while our intervention yields higher performance than both SN and the baseline.}
    \vspace{-0.25cm}
    \label{fig:sn_int}
\end{figure*}

\subsection{Regen}
\label{app:regen}

Following the results reported in~\cite{obando2026simplicial}, we conducted our experiments using ReGen with $\tau = 10^{-5}$. Specifically, we applied ReGen by zeroing weights whose magnitude falls below the threshold, thereby inducing sparsity in the network, following the procedure described by ~\cite{kumar2023maintaining}.

\begin{figure*}[!h]
    \centering
    \includegraphics[width=\textwidth]
    {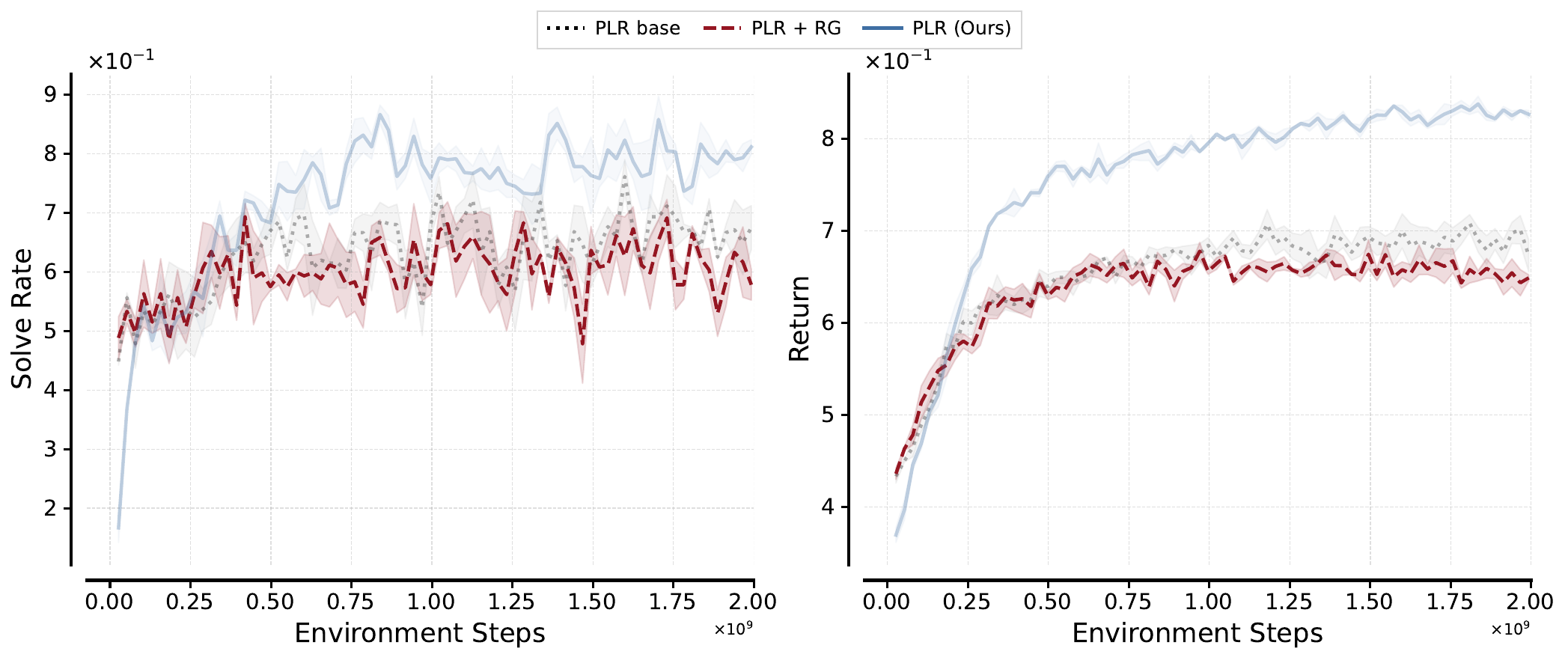}
    \vspace{-0.6cm}
    \caption{\textbf{Solve rate (left) and sampled return (right) over 5 seeds.}
    We compare the PLR baseline, ReGen, and the WC + CL (Ours) intervention. Our intervention consistently outperforms both the baseline and ReGen, achieving faster learning and higher values across training, while ReGen performs similarly to or below the baseline.}
    \vspace{-0.25cm}
    \label{fig:rg_int}
\end{figure*}

\subsection{HL-Gauss}
\label{app:categorical_value}

We use HL-Gauss to represent value predictions as a categorical distribution and train the critic using a cross-entropy loss. We evaluate this intervention with different numbers of bins: $21$, $51$, $101$, and $201$. This hyperparameter controls the resolution of the categorical value representation, and we compare each configuration against the PLR baseline ~\citep{farebrother2024stop}.

\begin{figure*}[!h]
    \centering
    \includegraphics[width=\textwidth]
    {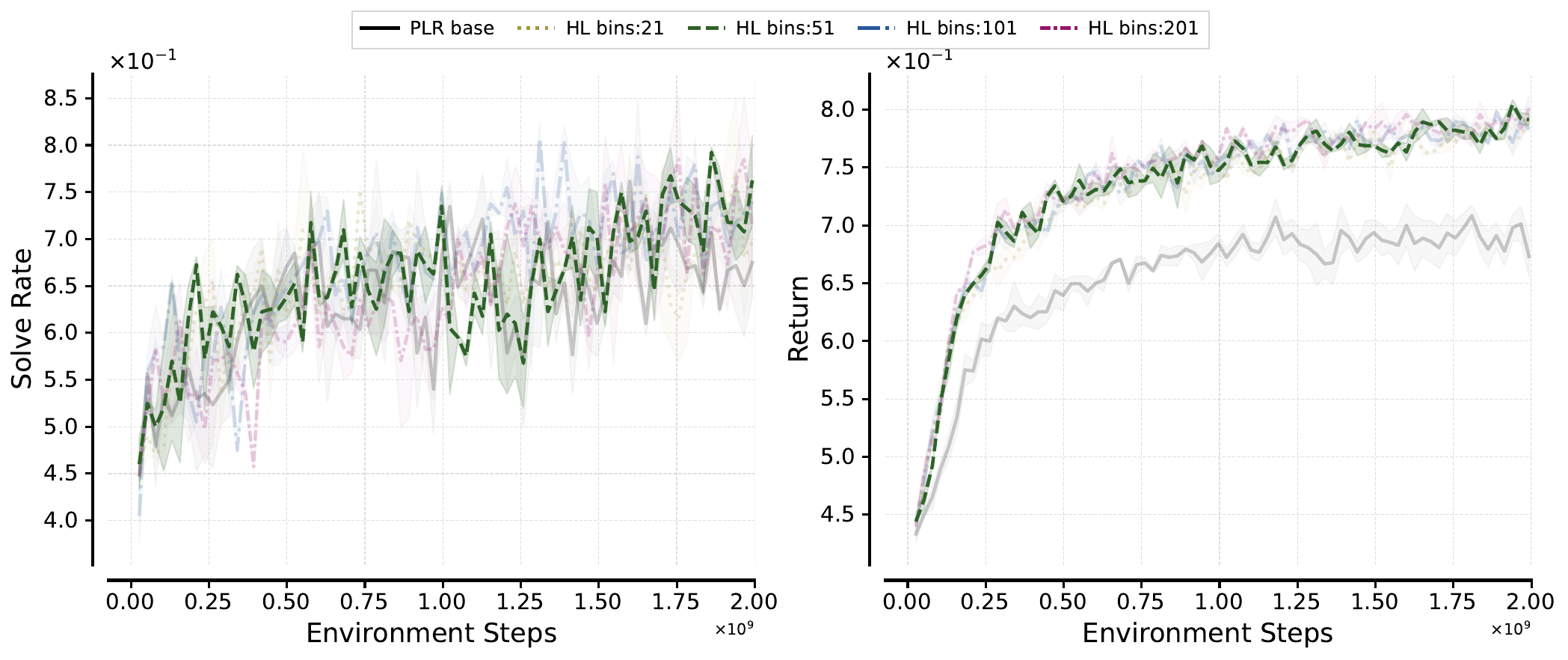}
    \vspace{-0.6cm}
    \caption{\textbf{Solve rate (left) and sampled return (right) over 5 seeds. }We compare the HL intervention on PLR using different numbers of bins: $21$, $51$, $101$, and $201$, which control the resolution of the categorical value representation. All tested configurations achieve similar results, consistent with the robustness to this hyperparameter reported in~\cite{farebrother2024stop}.}
    \vspace{-0.25cm}
    \label{fig:hl_test_int}
\end{figure*}

\begin{figure*}[!h]
    \centering
    \includegraphics[width=\textwidth]
    {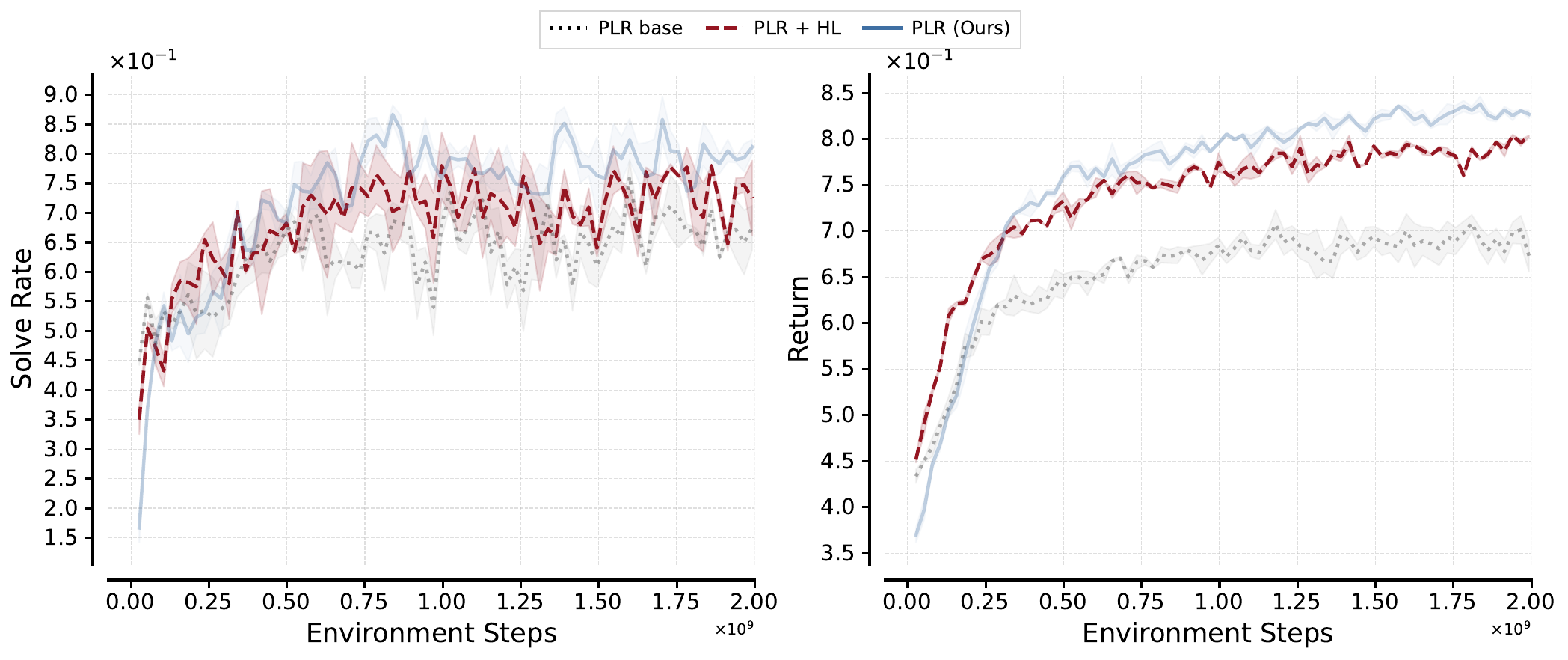}
    \vspace{-0.6cm}
    \caption{\textbf{Solve rate (left) and sampled return (right) over 5 seeds. }We consider the PLR baseline, HL with 51 bins, and our selected intervention. HL outperforms the PLR baseline, while our intervention performs better than both HL and the baseline.}
    \vspace{-0.25cm}
    \label{fig:hl_int}
\end{figure*}

\clearpage
\newpage

\section{Metrics}
\label{app:metrics}

\subsection{Solve Rate}
\label{app:solve_rate}

Solve Rate measures the proportion of evaluation episodes in which the agent successfully completes the task. We evaluate this metric on a static set of hand-designed physics-based holdout levels that are not used for training, providing a zero-shot evaluation of the agent's ability to generalize to unseen environments. A progressive increase in solve rate during training indicates that the agent is increasingly able to successfully solve these unseen tasks, reflecting improved generalization to physics-based challenges ~\citep{matthewskinetix}.

\subsection{Return Sampled}
\label{app:return_sampled}

Sampled Return measures the total return obtained by the agent over episodes evaluated on procedurally generated levels sampled from the Domain Randomization (DR) distribution. An increasing and sustained trend indicates that the agent is progressively obtaining higher returns across the sampled task distribution, suggesting improved performance across the diverse physics-based environments encountered during training ~\citep{matthewskinetix}.

\subsection{PVL}
\label{app:dr_pvl}

Positive Value Loss (PVL) is a heuristic metric used in Unsupervised Environment Design (UED) algorithms, such as PLR and ACCEL, to approximate regret and prioritize levels for replay. By considering only positive value losses, PVL selectively captures positive advantage estimates, indicating that the observed returns exceed those predicted by the critic. While a high PVL makes an individual level a higher-priority candidate for replay, when tracking PVL across the replay buffer, we expect it to be initially high and gradually decrease over time. This decline indicates that the agent's value estimates are becoming increasingly accurate and that the learning opportunities associated with these levels are being progressively resolved ~\citep{jiang2021replay}. 

\subsection{MaxMac}
\label{app:dr_maxmac}

Maximum Monte Carlo (MaxMC) is a heuristic scoring function used in UED algorithms such as PLR and ACCEL to approximate regret and prioritize levels for replay. Unlike PVL, which relies on the critic's value estimates, MaxMC uses the highest return previously achieved on a given level, denoted by $R_{\mathrm{max}}$. By replacing the value target with this maximum historical return, MaxMC reduces the dependence of the regret estimate on the agent's current policy and its value function. For an individual level, a high MaxMC score indicates a large gap between the best return previously achieved on that level and the critic's current value estimate, suggesting greater estimated learning potential. When tracking MaxMC for levels sampled through domain randomization (DR), a gradual increase indicates that the generated levels progressively exhibit higher maximum achieved returns, whereas abrupt changes may indicate sudden shifts in the properties or difficulty of the sampled levels ~\citep{jiang2021replay}.

\subsection{Value Prediction Mean}
\label{app:value_prediction_mean}

Mean Value Prediction measures the arithmetic mean of the critic's value estimates $V(s)$ across the states visited in a rollout batch, reflecting the expected future return estimated by the critic for the states encountered by the agent. A gradual increase in this metric during training can indicate that the agent is increasingly visiting states associated with higher expected returns, consistent with improved task performance. However, this metric should be interpreted alongside performance measures such as solve rate, since higher value estimates alone do not necessarily imply successful task completion ~\citep{schulman2017proximal}.  

\subsection{Gradient Norm}
\label{app:gradient_norm}

The gradient norm is the global $L_2$ norm of the gradient vector of the loss function with respect to all trainable parameters of the neural network. It indicates how strongly the optimizer needs to adjust the model parameters to correct value prediction errors and improve the policy. A stable and bounded gradient norm suggests that gradients propagate appropriately through the network and that learning progresses smoothly and continuously. It is computed as

$$
\text{grad\_norm}
=
\left\|\nabla_{\theta}\mathcal{L}\right\|_2
=
\sqrt{\sum_i
\left(
\frac{\partial \mathcal{L}}{\partial \theta_i}
\right)^2},
$$

where $\mathcal{L}$ denotes the loss function and $\theta_i$ represents the trainable parameters of the neural network.

\subsection{Representation Similarity}
\label{app:representation_similarity}

Representation Similarity measures the average cosine similarity between all latent representations (state embeddings) generated by the neural network within a training batch. Values close to 1 indicate that the representations point in similar directions in the latent space, suggesting a potential collapse of the learned representations. In contrast, values close to 0 indicate that the representations of different states are largely dissimilar and tend to point in orthogonal directions in the latent space.

\subsection{Spearman Correlation}
\label{app:spearman_correlation}

Spearman correlation measures the ranking fidelity of the Critic network. It evaluates whether the Critic can correctly rank the states within a rollout from better to worse by comparing its value predictions $V(s)$ with the corresponding empirical returns or targets $G_t$. A value close to 1 indicates that the Critic preserves the same ordering of states as reflected by their observed returns, suggesting high ranking fidelity.

\subsection{Target Mean}

Mean Value Target measures the arithmetic mean of the value targets computed from the trajectories collected in a rollout, using Generalized Advantage Estimation (GAE). It reflects the average return observed by the agent across the states encountered during the rollout. A gradual increase in this metric during training indicates that the agent is experiencing higher returns in the environments it visits, suggesting that it is progressively learning to achieve more rewarding outcomes ~\citep{schulman2017proximal}.

\subsection{Effective Rank}
\label{app:effective_rank}

This metric quantifies the dimensionality of learned representations by determining the smallest subspace that preserves at least 99\%  of the total variance. A higher feature rank indicates greater diversity in the learned representations. Following the approximate rank used by \cite{yang2019harnessing}, the feature rank $k$ is defined as the smallest number of singular values required to satisfy
\begin{equation}
\frac{\sum_{i=1}^{k}\sigma_i^2}
{\sum_{j=1}^{n}\sigma_j^2}
\geq \tau,
\end{equation}
where $\sigma_i$ denotes the $i$-th singular value of the feature matrix, $n$ is the total number of singular values, and $\tau$ is the variance threshold, set to $0.99$ in our experiments.

\subsection{Weight Magnitude}
\label{app:weight_magnitude}

This metric quantifies the magnitude of the neural network parameters, providing an indication of the scale of the learned weights. Larger weight norms correspond to parameters with greater magnitudes, which can affect the stability and generalization of the model. Following the formulation used by ~\cite{lyle2022understanding}, the weight norm is computed as
\begin{equation}
\lVert\theta\rVert_2 =\sqrt{\sum_i \theta_i^2},
\end{equation}
where $\theta_i$ denotes the weight parameters of a given layer.

\subsection{Number of Joints}
\label{app:number_of_joints}

In Kinetix, joints connect rigid bodies by imposing constraints that prevent the connected bodies from separating at the point of attachment. Fixed joints maintain a fixed relative configuration, whereas revolute joints allow relative rotation and can be actuated by motors that apply torque. The environment complexity increases with the maximum number of joints: Small (S) levels allow up to $1$ joint, Medium (M) levels up to $2$, and Large (L) levels up to $6$~\citep{matthewskinetix}. These joints increase the degrees of freedom of the system and can form coupled kinematic chains, where forces and torques applied at one joint can influence the dynamics of the connected bodies.

\subsection{Number of Shapes}
\label{app:number_of_shapes}

In Kinetix, the number of shapes determines the number of geometric components used to construct rigid bodies within the 2D scene. Together with Jax2D, Kinetix uses polygons and circles to represent the geometry of its physics-based environments~\citep{matthewskinetix}. The maximum number of shapes increases with environment complexity: Small (S) allows up to $7$ shapes, including at most $5$ polygons and $2$ circles; Medium (M) allows up to $9$ shapes, with at most $6$ polygons and $3$ circles; and Large (L) allows up to $16$ shapes, with at most $12$ polygons and $4$ circles.

\subsection{Number of Thrusters}
\label{app:number_of_thrusters}

The number of thrusters measures the maximum number of propulsion mechanisms present in a level. Each thruster provides an external linear propulsion force in the direction it points, giving the agent direct control over the motion of the associated body. The maximum number of thrusters increases with difficulty: Small (S) allows up to $1$ thruster, while Medium (M) and Large (L) allow up to $2$. This provides a range of task configurations, from environments that rely primarily on direct propulsion to more mechanically structured environments involving articulated bodies and joints~\citep{matthewskinetix}.

\subsection{Episode Length}
\label{app:episode_lenght}

Episode Length measures the average number of steps per episode, determined by whether the episode terminates after success, failure, or reaching the maximum episode horizon. We report this metric on the static, hand-designed holdout set, which is not used for training, as well as on levels sampled from the stochastic distribution used for Domain Randomization (DR). We refer to the latter as \textit{Episode Length (Sampled)} and report it separately for each environment size. Lower episode lengths indicate that episodes terminate in fewer steps, which, when associated with successful completion, reflects faster task completion ~\citep{matthewskinetix}.

\subsection{Learnability}
\label{app:learnability}

Learnability measures the immediate learning potential of a task or level given the agent's current capabilities. In PLR-based UED, this metric can be evaluated over either the curated replay buffer or the uncurated Domain Randomization (DR) stream ~\citep{matthewskinetix}. Higher learnability indicates that the sampled tasks lie closer to the agent's active learning region. For a level with success rate $p$, learnability is defined as
\begin{equation}
\mathcal{L} = p(1-p),
\end{equation}
where
\begin{equation}
p = \frac{N_{\mathrm{success}}}{N_{\mathrm{episodes}}}.
\end{equation}
For multiple environments, the aggregate learnability is computed as
\begin{equation}
\mathcal{L} = \sum_{e=1}^{N_{\mathrm{env}}} p_e(1-p_e).
\end{equation}
The metric is maximized at $p=0.5$ and becomes zero when $p=0$ or $p=1$, emphasizing levels that are neither too difficult nor too easy ~\citep{foster2026lilo}.   

\clearpage
\newpage

\section{Interventions Table}
\label{app:table_interventions}

Deep RL performance can be sensitive to experimental choices,
including hyperparameter selection, batch size, learning rates,
and network architecture
\citep{obando2023small,willi2024mixture,ceron2024on,
pmlr-v235-obando-ceron24b,pmlr-v235-obando-ceron24a,sokar2025don,obando2026simplicial,
schwarzer2023bigger}.
These choices are particularly relevant when evaluating learner
stability, as variations in training configurations can affect
both optimization dynamics and final performance.
We therefore evaluate stabilization interventions while holding
the curriculum rule, network architecture, and
environment-interaction budget fixed within each comparison.
This controlled experimental design supports more interpretable
and reproducible comparisons, consistent with broader efforts
toward rigorous empirical evaluation in deep RL
\citep{ceron2021revisiting}.

\begin{table}[!h]
\caption{Default PPO hyperparameter settings for Kinetix}
\label{humanoid-table}
\begin{center}
\begin{tabular}{ll}
\multicolumn{1}{c}{\bf Hyper-parameter}  &\multicolumn{1}{c}{\bf Value}
\\ \hline \\
Num Environments & 2048\\
Num Minibatches &  32\\
Num epochs & 8 \\
Learning Rate & 5e-5 \\
Discount Factor & 0.99\\
Gae Lambda & 0.9 \\
Vf Coef & 0.5\\
Entropy Coef & 0.01\\
Level Filter n Steps & 64 \\
Level Filter Sample Ratio & 2 \\
\end{tabular}
\end{center}
\end{table}

\begin{table}[!h]
\caption{Default ACCEL hyperparameter settings for Kinetix}
\label{accel-table}
\begin{center}
\begin{tabular}{ll}
\multicolumn{1}{c}{\bf Hyper-parameter} & \multicolumn{1}{c}{\bf Value}
\\ \hline \\
Use ACCEL & True \\
Exploratory Grad Updates & True \\
Num Edits & 5 \\
Score Function & MaxMC \\
Level Buffer Capacity & 4000 \\
Replay Probability & 0.5 \\
Staleness Coefficient & 0.3 \\
Temperature & 1.0 \\
Top-$k$ & 8 \\
Minimum Fill Ratio & 0.5 \\
Prioritization & Rank \\
\end{tabular}
\end{center}
\end{table}

\begin{table}[!h]
\caption{Default PLR hyperparameter settings for Kinetix}
%\label{accel-table}
\begin{center}
\begin{tabular}{ll}
\multicolumn{1}{c}{\bf Hyper-parameter} & \multicolumn{1}{c}{\bf Value}
\\ \hline \\
Use ACCEL & False \\
Exploratory Grad Updates & True \\
Num Edits & 2 \\
Score Function & MaxMC \\
Level Buffer Capacity & 4000 \\
Replay Probability & 0.5 \\
Staleness Coefficient & 0.3 \\
Temperature & 1.0 \\
Top-$k$ & 8 \\
Minimum Fill Ratio & 0.5 \\
Prioritization & Rank \\
\end{tabular}
\end{center}
\end{table}

\begin{table}[!h]
\caption{Default SFL hyperparameter settings for Kinetix}
\label{sfl-table}
\begin{center}
\begin{tabular}{ll}
\multicolumn{1}{c}{\bf Hyper-parameter} & \multicolumn{1}{c}{\bf Value}
\\ \hline \\
Sampled Environments Ratio & 0.5 \\
Batch Size & 4096 \\
Num Batches & 3 \\
Rollout Steps & 512 \\
Num to Save & 1024 \\
Buffer Update Frequency & 128 \\
\end{tabular}
\end{center}
\end{table}

\begin{table}[!h]
\caption{Hyperparameter settings for the evaluated interventions.}
\label{interventions-table}
\begin{center}
\begin{tabular}{lcc}
\multicolumn{1}{c}{\bf Intervention} &
\multicolumn{1}{c}{\bf Weight Clipping} &
\multicolumn{1}{c}{\bf Num. Bins} \\
\hline \\
Baseline & -- & -- \\
Weight Clipping (WC) & 2 & -- \\
HL-Gauss (HL) & -- & 51 \\
WC + HL & 2 & 51 \\
\end{tabular}
\end{center}
\end{table}

\begin{table}[!h]
\caption{Interventions evaluated in our experiments and their corresponding acronyms.}
\label{interventions-acronyms}
\begin{center}
\begin{tabular}{ll}
\multicolumn{1}{c}{\bf Acronym} & \multicolumn{1}{c}{\bf Intervention} \\
\hline \\
HL & HL-Gauss \\
RG & ReGen \\
SN-A & Spectral Normalization in the Actor \\
SN-B & Spectral Normalization in Both Actor and Critic \\
SN-C & Spectral Normalization in the Critic \\
WD & Weight Decay \\
WC & Weight Clipping \\
\end{tabular}
\end{center}
\label{app:acron}
\end{table}

\clearpage
\newpage

\section{Deep reinforcement learning}
\label{app:deep_rl}

We consider an agent interacting with a discounted Markov decision process
(MDP) $\mathcal{M}=(\mathcal{S},\mathcal{A},P,r,\gamma)$, where
$\mathcal{S}$ and $\mathcal{A}$ denote the state and action spaces,
$P(s' \mid s,a)$ is the transition kernel, $r(s,a)$ is the reward function,
and $\gamma\in[0,1)$ is the discount factor. At each timestep $t$, the agent
observes a state $s_t\in\mathcal{S}$, samples an action
$a_t\sim\pi_\theta(\cdot\mid s_t)$ from a parameterized policy
$\pi_\theta$, receives reward $r_t$, and transitions to
$s_{t+1}\sim P(\cdot\mid s_t,a_t)$. The objective is to maximize the expected
discounted return
\begin{equation}
    J(\theta)
    =
    \mathbb{E}_{\pi_\theta}
    \left[
        \sum_{t=0}^{\infty}\gamma^t r_t
    \right].
\end{equation}
The corresponding state-value and action-value functions are
\begin{align}
    V^{\pi_\theta}(s)
    &=
    \mathbb{E}_{\pi_\theta}
    \left[
        \sum_{t=0}^{\infty}\gamma^t r_t
        \,\middle|\,
        s_0=s
    \right], \\
    Q^{\pi_\theta}(s,a)
    &=
    \mathbb{E}_{\pi_\theta}
    \left[
        \sum_{t=0}^{\infty}\gamma^t r_t
        \,\middle|\,
        s_0=s,\,
        a_0=a
    \right],
\end{align}
with advantage
$A^{\pi_\theta}(s,a)=Q^{\pi_\theta}(s,a)-V^{\pi_\theta}(s)$
\citep{sutton2018reinforcement}.

We focus on actor--critic methods, which jointly optimize a policy
$\pi_\theta$ and a learned value function $V_\phi$. In Proximal Policy
Optimization (PPO)~\citep{schulman2017proximal}, trajectories are collected
under the current behavior policy $\pi_{\theta_{\mathrm{old}}}$ and reused
for multiple minibatch optimization steps. Defining the probability ratio
\begin{equation}
    \rho_t(\theta)
    =
    \frac{\pi_\theta(a_t\mid s_t)}
         {\pi_{\theta_{\mathrm{old}}}(a_t\mid s_t)},
\end{equation}
the clipped policy objective is
\begin{equation}
    \mathcal{L}_{\pi}(\theta)
    =
    -
    \mathbb{E}_{t}
    \left[
        \min\left(
            \rho_t(\theta)\widehat{A}_t,\,
            \operatorname{clip}
            \left(
                \rho_t(\theta),
                1-\epsilon,
                1+\epsilon
            \right)
            \widehat{A}_t
        \right)
    \right],
\end{equation}
where $\widehat{A}_t$ is an estimate of the advantage and $\epsilon$ controls
the clipping range. The critic is trained against trajectory-derived value
targets $\widehat{V}_t$, for example using the squared-error objective
\begin{equation}
    \mathcal{L}_{V}(\phi)
    =
    \mathbb{E}_{t}
    \left[
        \left(
            V_\phi(s_t)-\widehat{V}_t
        \right)^2
    \right].
\end{equation}

The actor and critic may use separate networks or share a learned
representation. In the shared setting, both policy and value losses update
the same features, coupling representation learning to both policy
optimization and value prediction.

Deep actor--critic learning is inherently non-stationary. At optimization
iteration $k$, the current policy $\pi_{\theta_k}$ induces a state-action
occupancy distribution $d^{\pi_{\theta_k}}(s,a)$. Updating the policy changes
this distribution and therefore changes the states and actions encountered
in subsequent trajectories. At the same time, updates to the value function
change the estimates used to construct value targets and advantages.
Consequently, the learner is repeatedly optimized on changing input
distributions and changing learning targets.

These dynamics can make sustained optimization difficult. In PPO,
representation rank deterioration and loss of capacity have been associated
with stronger policy-induced non-stationarity and, in some settings, eventual
performance collapse \citep{moalla2024no}. The way on-policy data are
collected also affects network plasticity and optimization stability, with
larger batches of independently collected experience often improving
learning behavior \citep{mayor2025impact}. Separately, learning plateaus in
PPO can arise when finite-sample estimates of the update objective become
poor proxies for the underlying objective, highlighting an additional source
of optimization stagnation in long-running on-policy training
\citep{beukman2026preventinglearningstagnationppo}. Together, these results
motivate studying both policy performance and the underlying optimization and
representation dynamics throughout training.

\section{Unsupervised environment design}
\label{app:ued}

Unsupervised environment design (UED) extends the standard RL setting from a
single MDP to a space of environment instances or levels
$\mathcal{L}$, where each level $\ell\in\mathcal{L}$ induces an MDP
$\mathcal{M}_{\ell}$ \citep{dennis2020emergent}. Instead of fixing the
training distribution over environments in advance, UED methods determine
which levels the student encounters during training.

At training iteration $k$, let $p_k(\ell)$ denote the distribution over
levels induced by the current curriculum mechanism. The student is trained
to improve its expected return under this distribution,
\begin{equation}
    J_k(\theta)
    =
    \mathbb{E}_{\ell\sim p_k}
    \left[
        J(\pi_\theta;\mathcal{M}_{\ell})
    \right].
\end{equation}
When the curriculum adapts over training, $p_k$ changes as a function of the
student's learning history, so the distribution of environments defining the
student's optimization problem also evolves over time.

UED methods differ in how they construct and update this curriculum.
Domain-randomization baselines sample levels from a fixed generator
distribution, whereas adaptive methods modify the level distribution using
information gathered during training. Prioritized Level Replay (PLR), for
example, preferentially revisits levels according to estimates of their
learning potential \citep{jiang2021prioritized}. Replay-guided adversarial
environment design combines replay-based level selection with environment
generation and places PLR within a broader UED framework
\citep{jiang2021replay}. Other approaches evolve environments toward the
frontier of the student's capabilities
\citep{parkerholder2022evolving}, while more recent work directly prioritizes
levels according to estimated learnability
\citep{rutherford2024no}. Thus, the curriculum need not itself be a
learned agent; it may instead be implemented through level generation,
scoring, replay, sampling, search, or evolutionary mechanisms.

Let $d_{\ell}^{\pi_{\theta_k}}(s,a)$ denote the state-action occupancy induced
by the current policy $\pi_{\theta_k}$ in level $\ell$. The effective
experience distribution at iteration $k$ can then be written as
\begin{equation}
    d_k(s,a)
    =
    \mathbb{E}_{\ell\sim p_k}
    \left[
        d_{\ell}^{\pi_{\theta_k}}(s,a)
    \right].
\end{equation}
This distribution can change through two mechanisms. First, even for a fixed
level distribution, policy updates change the occupancy
$d_{\ell}^{\pi_{\theta_k}}$. Second, under an adaptive curriculum, updates to
$p_k$ change the mixture of environments from which experience is collected.
Adaptive UED therefore introduces curriculum-induced distribution shift in
addition to the policy-induced non-stationarity already present in standard
deep RL.

More generally, if $\mathcal{D}_k$ denotes trajectories collected by sampling
$\ell\sim p_k$ and executing the current policy, the learner update can be
written abstractly as
\begin{equation}
    w_{k+1}
    =
    \mathcal{U}(w_k;\mathcal{D}_k),
\end{equation}
where $w_k$ contains the trainable parameters of the student. For an adaptive
curriculum, we can similarly write
\begin{equation}
    p_{k+1}
    =
    \mathcal{C}(p_k,\mathcal{H}_{k+1}),
\end{equation}
where $\mathcal{H}_{k+1}$ denotes the information used by the curriculum,
such as student trajectories, returns, level scores, or replay statistics.
The learner and curriculum are therefore coupled: the current curriculum
determines the experience used to update the student, while the student's
behavior determines the information available for constructing subsequent
curricula.

This coupling creates a particularly non-stationary learning problem.
Changes in the curriculum alter the mixture of states, transitions, rewards,
and task difficulties encountered during training, while policy updates
simultaneously alter visitation within each level. The critic must therefore
track value targets under an evolving training distribution, and the actor
must optimize advantage estimates computed from experience generated by both
a changing policy and, for adaptive methods, a changing curriculum. When the
actor and critic share a representation, these shifts also continually
change the feature-learning problem faced by the network.

We use this setting to study how curriculum-induced distribution shift
interacts with the optimization and representation dynamics of deep RL
agents, and in particular whether mechanisms that improve optimization
stability help the student continue adapting as the training distribution
evolves.

\clearpage
\newpage

\section{Experiments}
\label{app:experiments}

\begin{figure*}[!h]
    \centering
   \includegraphics[width=0.8\linewidth]
        {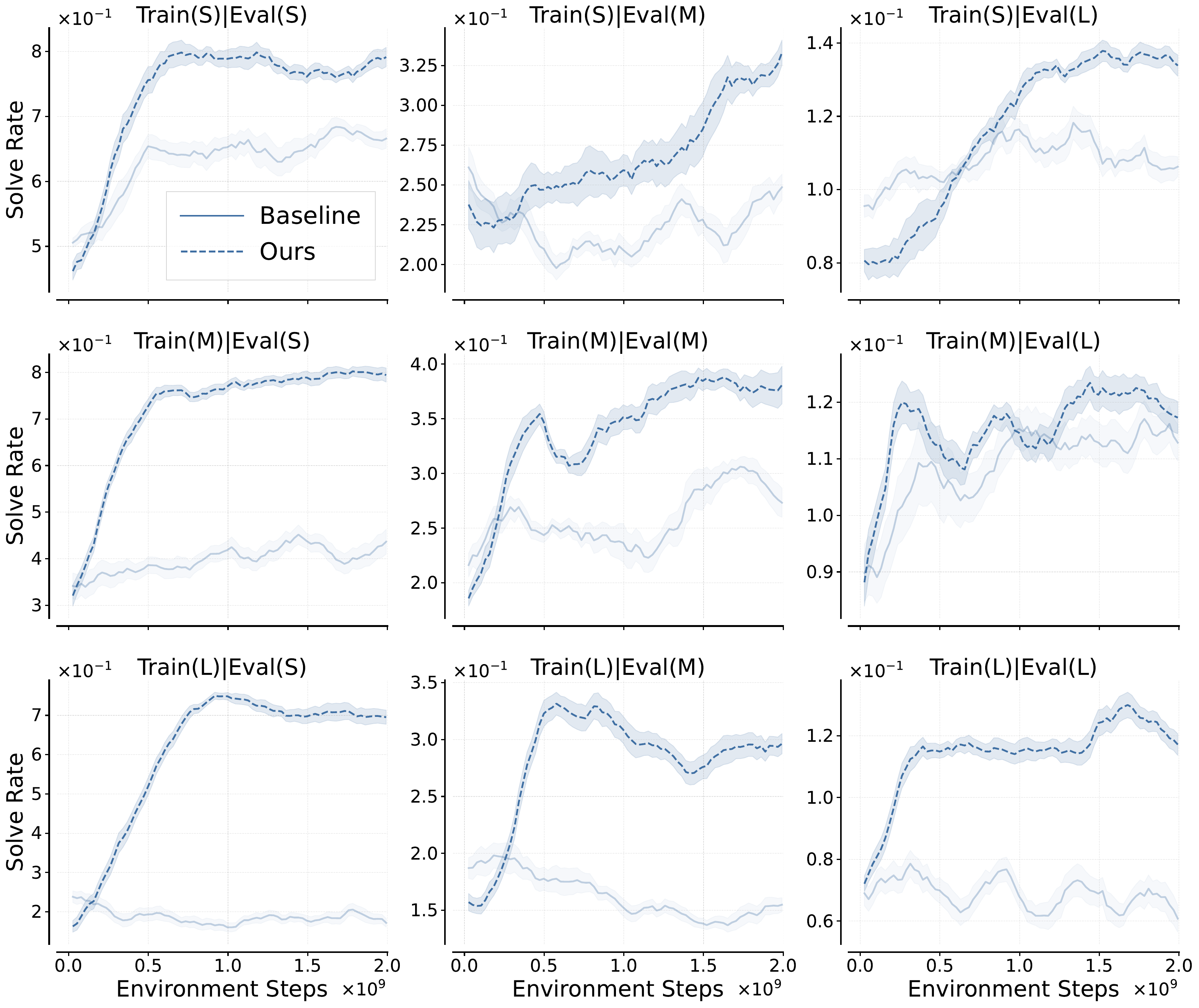}
        \captionof{figure}{\textbf{Generalization to human-designed holdout tasks.} Zero-shot solve rate across training and evaluation complexities.}
    \vspace{-0.25cm}
    \label{fig:heldout_cross_complexity}
\end{figure*}

\begin{figure*}[!h]
    \centering
           \includegraphics[width=0.8\linewidth]
        {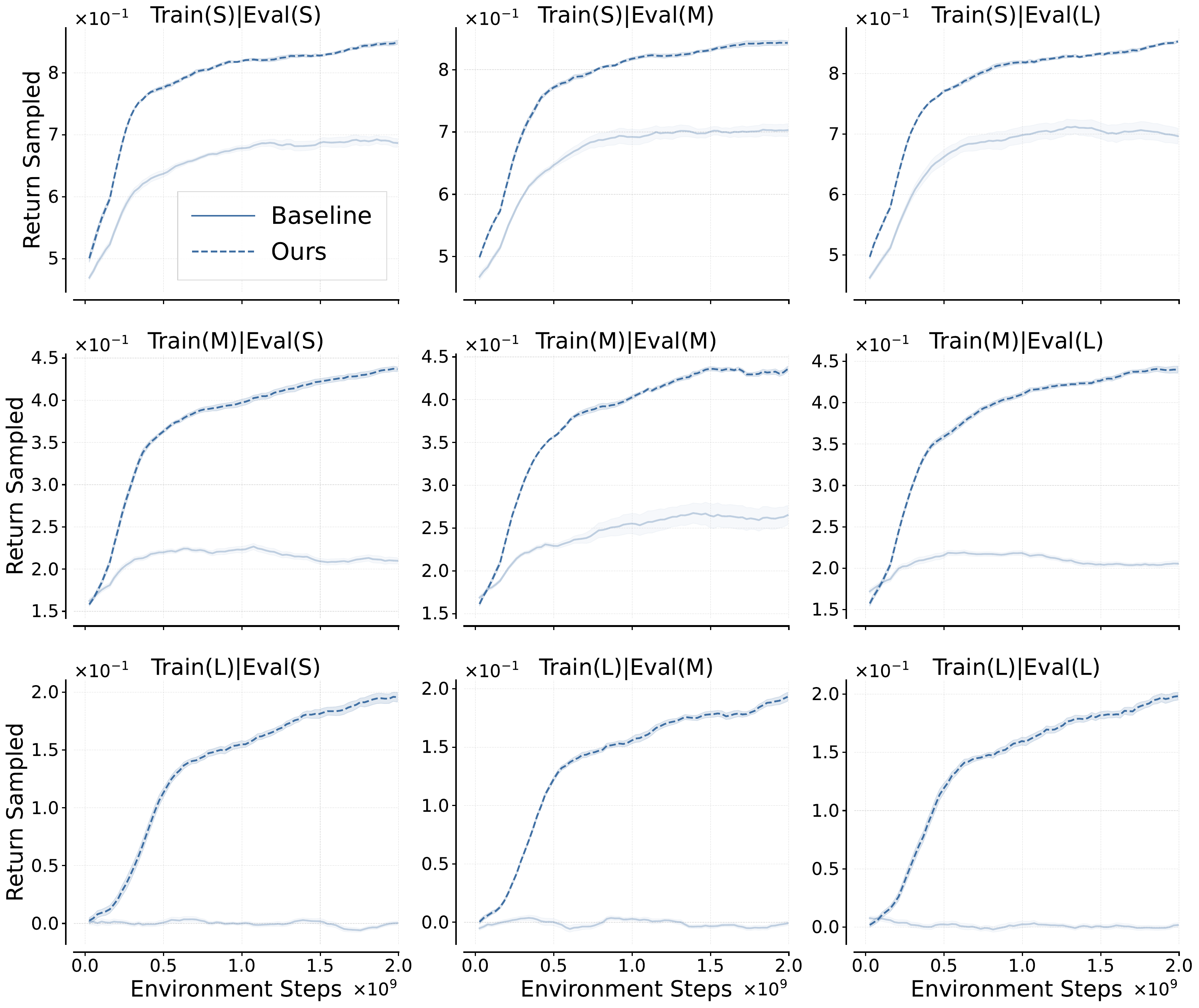}
        \captionof{figure}{
        \textbf{Generalization to randomly generated tasks.} Zero-shot sampled return across training and evaluation complexities.}
    \vspace{-0.25cm}
    \label{fig:sampled_cross_complexity}
\end{figure*}